\documentclass[11pt,3p,review,authoryear]{elsarticle}
\journal{Expert Systems with Applications}

\usepackage{url}
\usepackage{amsmath}
\usepackage{fancybox}
\usepackage{hyperref}
\usepackage{amsfonts}
\usepackage{amsmath}
\usepackage{multirow}
\usepackage[lined,boxruled,linesnumbered]{algorithm2e}
\usepackage{enumitem}
\usepackage{lscape}
\usepackage{rotating}
\usepackage{subfig}

\usepackage{import}

\usepackage{color}
\usepackage{pifont}
\usepackage{wasysym}
\usepackage{multirow}
\usepackage{array}

\definecolor{cblue}{rgb}{0.2,0.2,0.73}
\definecolor{cgray}{rgb}{0.72,0.72,0.77}
\definecolor{darkblue}{rgb}{0,0,0.55}
\definecolor{darkgreen}{rgb}{0,0.39,0}
\definecolor{darkred}{rgb}{0.55,0,0}
\definecolor{urlblue}{rgb}{0.3,0.43,0.7}
\definecolor{urlblue2}{rgb}{0,0,0.5}

\newcolumntype{M}[1]{>{\raggedright}m{#1}}

\begin{document}
\title{A review and comparison of strategies for multi-step ahead time series forecasting based on the NN5 forecasting competition}
\author[mlg]{Souhaib Ben Taieb}
\author[mlg]{Gianluca Bontempi}
\author[cairo]{Amir Atiya}
\author[hut]{Antti Sorjamaa}
\address[mlg]{Machine Learning Group, D\'epartement d'Informatique, Facult\'e des Sciences, Universit\'e Libre de Bruxelles, Belgium}
\address[hut]{Environmental and Industrial Machine Learning Group, Adaptive Informatics Research Centre, Altoo University School of Science, Finland}
\address[cairo]{Faculty of Engineering, Cairo University, Giza, Egypt}

\begin{abstract}
Multi-step ahead forecasting is still an open challenge in time series forecasting. 
Several approaches that deal with this complex problem have been
proposed in the literature but an extensive comparison on a large
number of tasks is still missing. This paper aims to fill this gap by reviewing existing strategies for multi-step ahead forecasting
and comparing them in theoretical and practical terms. To attain such an objective, we performed a large scale comparison of these different strategies using a large experimental 
benchmark (namely the 111 series from the NN5 forecasting competition).
In addition, we considered the effects of deseasonalization, input variable selection,
and forecast combination on these strategies and on multi-step ahead forecasting at large.
The following three findings appear to be consistently supported by the experimental results: Multiple-Output strategies are the best performing approaches, deseasonalization leads to uniformly improved forecast accuracy, and input selection is more effective when performed in conjunction with deseasonalization.

\end{abstract}

\maketitle

\clearpage

\thispagestyle{empty} 
\tableofcontents

\clearpage

\section{Introduction}

Time series forecasting is a growing field of interest playing an important role in nearly all fields of science and engineering, such as economics, finance, meteorology and telecommunication~\citep{compTimeSeries2005}. Unlike one-step ahead forecasting, multi-step ahead forecasting tasks are more difficult~\citep{tiaotsay}, since they have to deal with various additional complications, like accumulation of errors, reduced accuracy, and increased uncertainty~\citep{santafe,longterm}.

The forecasting domain has been influenced, for a long time, by linear statistical methods such as ARIMA models. However, in the late 1970s and early 1980s, it became increasingly clear that linear models are not adapted to many real applications~\citep{GH06}. In the same period, several useful nonlinear time series models were proposed such as the bilinear model~\citep{PT86}, the threshold autoregressive model~\citep{TongRS1980,Tong83,Tong00} and the autoregressive conditional heteroscedastic~(ARCH) model~\citep{Engle1982}~(see \citep{GH06}  and \citep{DeGooijer1992135} for a review). Nowadays, Monte Carlo simulation or bootstrapping methods are used to compute nonlinear forecasts. Since no assumptions are made about the distribution of the error process, the latter approach is preferred~\citep{CFS04,GH06}. However, the study of nonlinear time series analysis and forecasting is still in its infancy compared to the development of linear time series~\citep{GH06}.

In the last two decades, machine learning models have drawn attention and have established themselves as serious contenders to classical statistical models in the forecasting community~\citep{empiricalMLTimeSeries,compTimeSeries2005,ZPH98}. These models, also called black-box or data-driven models~\citep{mlmitchel}, are examples of nonparametric nonlinear models which use only historical data to learn the stochastic dependency between the past and the future. For instance, Werbos found that Artificial Neural Networks~(ANNs) outperforms the classical statistical methods such as linear regression and Box-Jenkins approaches~\citep{Werbos74,Werbos1988}. A similar study has been conducted by Lapedes and Farber~\citep{Lapedes87} who conclude that ANNs can be successfully used for modeling and forecasting nonlinear time series. Later, others models appeared such as decision trees, support vector machines and nearest neighbor regression~\citep{hastie_09_elements-of.statistical-learning,alpaydin-04-ml}. Moreover, the empirical accuracy of several machine learning models has been explored in a number of forecasting competitions under different data conditions~(e.g. the NN3, NN5, and the annual ESTSP competitions~\citep{nn3,nn5,estsp07,estsp08}) creating interesting scientific debates in the area of data mining and forecasting~\citep{debateMiningForecasting,debateMiningForecastingCommentPrice,debateMiningForecastingCommentCrone}.

In the forecasting community, researchers have paid attention to several aspects of the forecasting procedure such as model selection~\citep{Aha97,msejor06,msnn99,Chapelle00modelselection}, effect of deseasonalization~\citep{modellingSeas,fma,nelson99,Zhang2005501}, forecasts combination~\citep{Bates69thecombination,Clemen89,forecastcomb} and many other critical topics~\citep{GH06}. However, approaches for generating multi-step ahead forecasts for machine learning models did not receive as much attention, as pointed out by Kline: ``\emph{One issue that has had limited investigation is how to generate multiple-step-ahead forecasts}''~\citep{klineMethods}.

To the best of our knowledge, five alternatives~(or strategies) have been proposed in the literature to tackle an $H$-step ahead forecasting task. The \emph{Recursive} strategy~\citep{santafe, longterm,multistep,tiaotsay,klineMethods,Hamzacebi20093839} iterates, $H$ times, a one-step ahead forecasting model to obtain the $H$ forecasts. After estimating the future series value, it is fed back as an input for the following forecast. 

In contrast to the previous strategy which use a single model, the \emph{Direct} strategy~\citep{santafe, longterm,multistep,tiaotsay,klineMethods,Hamzacebi20093839} estimates a set of $H$ forecasting models, each returning a forecast for the $i$-th value~($i \in \{1,\dots,H\} $). 

A combination of the two previous strategies, called \emph{DirRec} strategy has been proposed in~\citep{sorjamaa_esann_06}. The idea behind this strategy is to combine aspects from both, the Direct and the Recursive strategies. In other words, a different model is used at each step but the approximations from previous steps are introduced into the input set. 

In order to preserve, between the predicted values, the stochastic dependency characterizing the time series, the Multi-Input Multi-Output~(\emph{MIMO}) strategy has been introduced and analyzed in ~\citep{mimo,mimoijf,klineMethods}. Unlike the previous strategies where the models return a scalar value, the MIMO strategy returns a vector of future values in a single step. 

The last strategy, called \emph{DIRMO}~\citep{mismo}, aims to preserve the most appealing aspects of both the DIRect and miMO strategies. This strategy aims to find a trade-off between the property of preserving the stochastic dependency between the forecasted values and the flexibility of the modeling procedure. 

In the literature, these five forecasting strategies have been presented separately, sometimes, using different terminologies. The \emph{first contribution} of this paper is to present a thorough unified review as well as a theoretical comparative analysis of the existing strategies for multi-step ahead forecasting.

Despite the fact that many studies have compared between the different multi-step ahead approaches, the collective outcome of these studies regarding forecasting performance has been inconclusive.  So the modeler is still left with little guidance as to which strategy to use. For example, research from \citep{Bontempi99BirattariBersini-ICML99,Weigend92HubermanRumelhart} provide experimental evidence in favor of Recursive strategy against Direct strategy. However, results from \citep{Zhang94Hutchinson,longterm,Hamzacebi20093839} support the fact that the Direct strategy is better than the Recursive strategy.  The work by \citep{sorjamaa_esann_06} shows that the DirRec strategy gives better performance than Direct and Recursive strategies. The Direct and Recursive strategies have been theoretically and empirically compared in \citep{Atiya99acomparison}. In this study the authors obtained theoretical and experimental evidence in favor of Direct strategy. Concerning the MIMO strategy, Kline~\citep{klineMethods} and Cheng et al~\citep{multistep} support the idea that MIMO strategy provides worse forecasting performance than Recursive and Direct strategies. However, in \citep{mimo,mimoijf}, the comparison between MIMO, Recursive, and Direct was in favor of MIMO. Finally, ~\citep{mismo,BenTaieb20101950} show that the DIRMO strategy gives better forecasting results than Direct and MIMO strategies when the parameter controlling the degree of dependency between forecasts is correctly identified.
These previous comparisons have been performed with different datasets in different configurations using different forecasting methods, such as Multiple Linear Regression, Artificial Neural Networks, Hidden Markov Models and Nearest Neighbors. 

All the contradictory findings of these studies make it all the more necessary to investigate further to find the truth concerning the relative performance of these strategies.
The \emph{second contribution} of this paper is an experimental comparison of the different multi-step ahead forecasting strategies on the $111$ time series of the NN5 international forecasting competition benchmark. These time series pose some of the realistic problems that one usually encounters in a typical multi-step ahead forecasting task, for example the existence of several times series of possibly related dynamics, outliers, missing values, and multiple overlying seasonalities. This experimental comparison is performed for a variety of different configurations (regarding seasonality, input selection and combination), in order to have the comparison as encompassing as can be. In addition, the methodology used for this experimental comparison is based on the guidelines and recommendations advocated in some of the methodological papers \citep{Demsar_statisticalcomparisons,extensionDemsar}.

In other words, the aim of this paper is not to make a comparison of machine learning algorithms for forecasting~(which was already conducted in \citep{empiricalMLTimeSeries}) but rather to show for a given learning algorithm, how the choice of the forecasting strategy can sensibly influence the performance of the multi-step ahead forecasts. In this work, we adopted the Lazy Learning algorithm~\citep{Aha97}, a particular instance of local learning, which has been successfully applied to many real-world forecasting tasks~\citep{Sauer94,Bontempi98BirattariBersini_Leuven,McNames98}. 

Last but not least, the paper proposes also a Lazy Learning entry to the NN5 forecasting competition~\citep{nn5}. The goal is to assess how this model fares compared to the other computational intelligence models that were proposed for the competition~\citep{mimoijf}. This will give us an idea about the potential of this approach.

The paper is organized as follows. The next section presents a review of the different forecasting strategies. Section \ref{sectionLazy} describes the Lazy Learning model and the associated algorithms for the different forecasting strategies. Section \ref{sectionExperiments} gives a detailed presentation of the datasets and the methodology applied for the experimental comparison. Section \ref{sectionResDiscussion} presents the results and discusses them. Finally, Section \ref{sectionConclusion} gives a summary and concludes the work.

\section{Strategies for Multi-Step-Ahead Time Series Forecasting}

\label{sectionStrategies}

A multi-step ahead~(also called long-term) time series forecasting task consists of predicting the next $H$ values $[y_{N+1},\dots,y_{N+H}]$ of a historical time series $[y_1,\dots,y_N]$ composed of $N$ observations, where $H>1$ denotes the forecasting horizon.

This section will first give a presentation of the five forecasting strategies and next, a subsection will be devoted to a comparative analysis of these strategies in terms of number and types of models to learn as well as forecasting properties.

We will use a common notation where $f$ and $F$ denote the functional dependency between past and future observations, $d$ refers to the embedding dimension~\citep{Casdagli91EubankFarmerGibson} of the time series, that is the number of past values used to predict future values and $w$ represents the term that includes modeling error, disturbances and/or noise.

\subsection{Recursive strategy}
\label{RecSection}

The oldest and most intuitive forecasting strategy is the \emph{Recursive}~(also called \emph{Iterated} or \emph{Multi-Stage}) strategy~\citep{santafe, longterm,multistep,tiaotsay,klineMethods,Hamzacebi20093839}. In this strategy, a single model $f$ is trained to perform a {\it one-step ahead} forecast, i.e.

\begin{equation} y_{t+1}=f(y_t,\dots,y_{t-d+1})+w \textnormal{,} \label{stochRec} \\ \end{equation}

\noindent with $t \in \{d,\dots,N-1\}$.

When forecasting $H$ steps ahead, we first forecast the first step
by applying the model. Subsequently, we use the value just forecasted
as part of the input variables for forecasting the next step (using the
same one-step ahead model).
We continue in this manner until we have forecasted
the entire horizon.

Let the trained one-step ahead model be $\hat f$. Then the forecasts
are given by:

\begin{equation}
\label{equationRec}
\hat y_{N+h} = 
\begin{cases} 
\hat{f}(y_{N},\dots,y_{N-d+1}) & \text{if } h=1 \\
\hat{f}(\hat y_{N+h-1}, \dots,\hat y_{N+1},y_{N},\dots,y_{N-d+h}) & \text{if } h \in \{2,\dots,d\} \\
\hat{f}(\hat y_{N+h-1}, \dots,\hat y_{N+h-d}) & \text{if } h \in \{d+1,\dots,H\} \\
\end{cases}
\end{equation}

Depending on the noise present in the time series and the forecasting horizon, the recursive strategy may suffer from low performance in multi-step ahead forecasting tasks. Indeed, this is especially true if the forecasting horizon $h$ exceeds the embedding dimension $d$, as at some point all the inputs are forecasted values instead of actual observations~(Equation \ref{equationRec}). The reason for the potential inaccuracy is that the Recursive strategy is sensitive to the accumulation of errors with the forecasting horizon. 
Errors present in intermediate forecasts will propagate forward as these forecasts
are used to determine subsequent forecasts.

In spite of these limitations, the Recursive strategy has been successfully used to forecast many real-world time series by using different machine learning models, like recurrent neural networks~\citep{rnnTimeSeries} and nearest-neighbors~\citep{McNames98,Bontempi99BirattariBersini-ICML99}.

\subsection{Direct strategy}

The \emph{Direct}~(also called Independent) strategy~\citep{santafe, longterm,multistep,tiaotsay,klineMethods,Hamzacebi20093839} consists of forecasting each horizon independently from the others. In other terms, $H$ models $f_h$ are learned~(one for each horizon) from the time series $[y_1,\dots,y_N]$ where

\begin{equation} y_{t+h}=f_h(y_t,\dots,y_{t-d+1})+w \label{stochDirect} \textnormal{,} \\ \end{equation}

with $t \in \{d,\dots,N-H\}$ and $h \in \{1,\dots,H\}$.

The forecasts are obtained by using the $H$ learned models $\hat{f}_h$ as follows:

\begin{equation}
\label{equationDir}
\hat{y}_{N+h}=\hat{f}_h(y_{N},\dots,y_{N-d+1}) \textnormal{.} 
\end{equation}

This implies that the Direct strategy does not use any approximated values to compute the forecasts~(Equation \ref{equationDir}), being then immune to the accumulation of errors. However, the $H$ models are learned independently inducing a conditional independence of the $H$ forecasts. This affects the forecasting accuracy as it prevents the strategy from considering complex dependencies between the variables $\hat{y}_{N+h}$\citep{mimo,mimoijf,klineMethods}. For example consider a case where the best forecast is a linear or mildly nonlinear trend. The direct method could yield a broken curve because of the ``uncooperative" way the $H$ forecasts are generated. Also, this strategy demands a large computational time since there are as many models to learn as the size of the horizon.

Different machine learning models have been used to implement the Direct strategy for multi-step ahead forecasting tasks, for instance neural networks~\citep{klineMethods}, nearest neighbors~\citep{longterm} and decision trees~\citep{directTrees}.

\subsection{DirRec strategy}

The \emph{DirRec} strategy~\citep{sorjamaa_esann_06} combines the architectures and the principles underlying the Direct and the Recursive strategies. DirRec computes the forecasts with different models for every horizon~(like the Direct strategy) and, at each time step, it enlarges the set of inputs by adding variables corresponding to the forecasts of the previous step~(like the Recursive strategy). However, note that unlike the two previous strategies, the embedding size $d$ is not the same for all the horizons. In other terms, the DirRec strategy learns $H$ models $f_h$ from the time series $[y_1,\dots,y_N]$ where

\begin{equation} y_{t+h}=f_h(y_{t+h-1},\dots,y_{t-d+1})+w \textnormal{,} \\  \label{stochDirRec} \end{equation}

\noindent with $t \in \{d,\dots,N-H\}$ and $h \in \{1,\dots,H\}$.

To obtain the forecasts, the $H$ learned models are used as follows:

\begin{equation}
\label{equationDirRec}
\hat y_{N+h} = 
\begin{cases} 
\hat{f_h}(y_{N},\dots,y_{N-d+1}) & \text{if } h=1 \\
\hat{f_h}(\hat y_{N+h-1}, \dots,\hat y_{N+1},y_{N},\dots,y_{N-d+1}) & \text{if } h \in \{2,\dots,H\} \\
\end{cases}
\end{equation}

This strategy outperformed the Direct and the Recursive strategies on two real-world time series: Santa Fe and Poland Electricity Load data sets~\citep{sorjamaa_esann_06}. Few research has been done regarding this strategy, so there is a need for further evaluation.
\subsection{MIMO strategy}

The three previous strategies~(Recursive, Direct and DirRec) may be considered as Single-Output strategies~\citep{BenTaieb20101950} since they model the data as a (multiple-input) single-output function~(see Equations \ref{equationRec}, \ref{equationDir} and \ref{equationDirRec}). 

The introduction of the \emph{Multi-Input Multi-Output}~(MIMO) strategy~\citep{mimo,mimoijf}~(also called Joint strategy~\citep{klineMethods}) has been motivated by the need to avoid the modeling of single-output mapping, which neglects the existence of stochastic dependencies between future values and consequently affects the forecast accuracy~\citep{mimo,mimoijf}. 

The MIMO strategy learns one multiple-output model $F$ from the time series $[y_1,\dots,y_N]$ where

\begin{equation}  [y_{t+H},\dots,y_{t+1}]=F(y_{t},\dots,y_{t-d+1})+ \mathbf{w} \textnormal{,} \\  \label{stochMIMO} \end{equation}

\noindent with $t \in \{d,\dots,N-H\}$, $F: \mathbb{R}^d \rightarrow \mathbb{R}^H$ is a vector-valued function~\citep{vector-valued}, and $\mathbf{w} \in \mathbb{R}^H$ is a noise vector with a covariance that is not necessarily diagonal~\citep{matias} .

The forecasts are returned in one step by a multiple-output model $\hat F$ where

\begin{equation} \label{equationMIMO} [\hat{y}_{t+H},\dots,\hat{y}_{t+1}]=\hat{F}(y_{N},\dots,y_{N-d+1}) \textnormal{.} \\ \end{equation}

The rationale of the MIMO strategy is to preserve, between the predicted values, the stochastic dependency characterizing the time series. This strategy avoids the conditional independence assumption made by the Direct strategy as well as the accumulation of errors  from which plagues the Recursive strategy. So far, this strategy has been successfully applied to several real-world multi-step ahead time series forecasting tasks~\citep{mimo,mimoijf,mismo,BenTaieb20101950}.

However, the need to preserve the stochastic dependencies by using one model has a drawback as it constrains all the horizons to be forecasted with the same model structure. This constraint could reduce the flexibility of the forecasting approach~\citep{mismo}. This was the motivation for the introduction of a new multiple-output strategy: DIRMO~\citep{mismo,BenTaieb20101950}, presented next.

\subsection{DIRMO strategy}

The DIRMO strategy~\citep{mismo} aims to preserve the most appealing aspects of both the DIRect and miMO strategies. Taking a middle approach, DIRMO forecasts the horizon $H$ in blocks, where each block is forecasted in a MIMO fashion. Thus, the $H$-step-ahead forecasting task is decomposed into $n$ multiple-output forecasting tasks~($n=\frac{H}{s}$), each with an output of size $s$~($s \in \{1,\dots,H\}$).

When the value of the parameter $s$ is $1$, the number of forecasting tasks $n$ is equal to $H$ which correspond to the Direct strategy. When the value of the parameter $s$ is $H$, the number of forecasting tasks $n$ is equal to $1$ which correspond to the MIMO strategy. There are intermediate configurations between these two extremes depending on the value of a parameter $s$. 

The tuning of the parameter $s$ allows us to improve the flexibility of the MIMO strategy by calibrating the dimensionality of the outputs~(no dependency in the case $s=1$ and maximal dependency for $s=H$). This provides a beneficial trade off between the preserving a larger degree of the stochastic dependency between future values and having a greater flexibility of the predictor.

The DIRMO strategy, previously called MISMO strategy~\citep{mismo}~(renamed for clarity reason), learns $n$ models $F_p$ from the time series $[y_1,\dots,y_N]$ where 

\begin{equation}  [y_{t+p*s},\dots,y_{t+(p-1)*s+1}]=F_p(y_{t},\dots,y_{t-d+1})+ \mathbf{w} \textnormal{,} \\ \label{stochDIRMO} \end{equation}

\noindent with $t \in \{d,\dots,N-H\}$, $p \in \{1,\dots,n\}$ and $F_p: \mathbb{R}^d \rightarrow \mathbb{R}^s$ is a vector-valued function if $s>1$.

The $H$ forecasts are returned by the $n$ learned models as follows:

\begin{equation} \label{equationDIRMO}  [\hat{y}_{N+p*s},\dots,\hat{y}_{N+(p-1)*s+1}]=\hat{F}_p(y_{N},\dots,y_{N-d+1}) \textnormal{.} \\ \end{equation}

The DIRMO strategy has been successfully applied to two forecasting competitions: ESTSP'07~\citep{mismo} and NN3~\citep{BenTaieb20101950}.  


\subsection{Comparative Analysis}

\begin{figure}[h]
\centering
\includegraphics[scale=0.8]{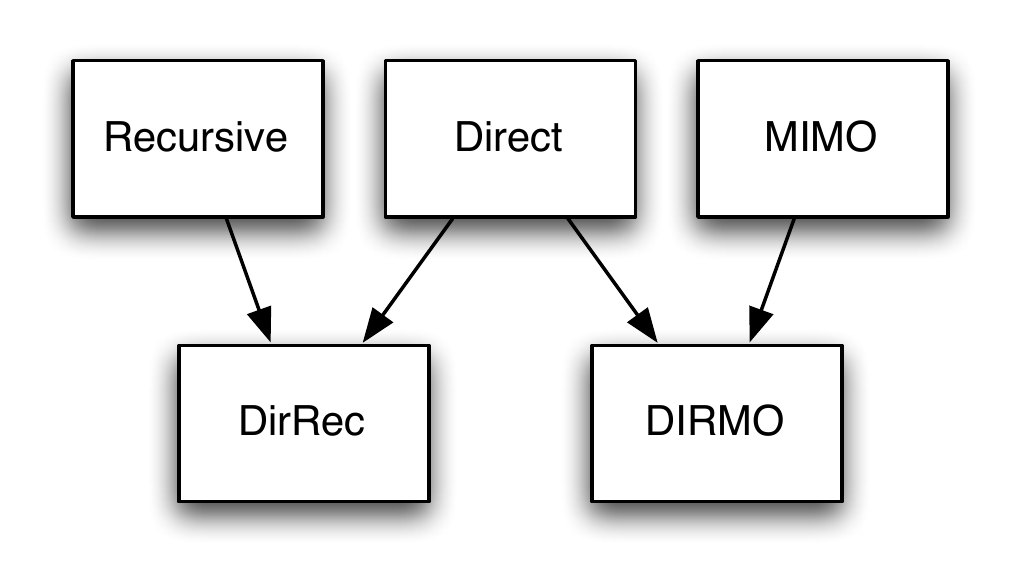}
\caption{\label{strategiesGraph} The different forecasting strategies with the links showing their relationship.}
\end{figure}

To summarize, there are five possible forecasting strategies that perform a multi-step ahead forecasting task: \emph{Recursive}, \emph{Direct}, \emph{DirRec}, \emph{MIMO} and \emph{DIRMO} strategies. Figure \ref{strategiesGraph} shows the different forecasting strategies with links indicating their relationships.

As we see, the DirRec strategy is a combination of the Direct and the Recursive strategy, while the DIRMO strategy is a combination of the Direct and the MIMO strategy.

Contingent on the selected strategy, a different number and type of models will be required. Before presenting the general comparison of the multi-step ahead forecasting strategies, let us highlight using an example the differences between the forecasting strategies.

Consider a multi-step ahead forecasting task for the time series $[y_1,\dots,y_N]$ where the forecasting horizon $H$ is $4$. Table \ref{exampleStrategies} shows, for each strategy, the different input sets and forecasting models involved in the calculation of the four forecasts $[\hat{y}_{N+1},\dots,\hat{y}_{N+4}]$.

\begin{table}[h]
\begin{tabular}{|l|l|l|l|l|}
\hline
&   $   \hat{y}_{N+1}$ & $\hat{y}_{N+2}$ & $\hat{y}_{N+3}$ & $\hat{y}_{N+4}   $ \\
  \hline
  \hline
Recursive &      $ {\scriptstyle   \hat{f}(y_{N},\dots,y_{N-d+1})   } $  & ${\scriptstyle    \hat{f}(\hat{y}_{N+1},y_{N},\dots,y_{N-d+2})   }$   & ${\scriptstyle   \hat{f}(\hat{y}_{N+2},\hat{y}_{N+1},\dots,y_{N-d+3})   }$ & ${\scriptstyle   \hat{f}(\hat{y}_{N+3},\hat{y}_{N+2},\dots,y_{N-d+4})  }$  \\
    \hline
Direct &      ${\scriptstyle  \hat{f}_1(y_{N},\dots,y_{N-d+1})  }$ & ${\scriptstyle   \hat{f}_2(y_{N},\dots,y_{N-d+1})  }$ & ${\scriptstyle  \hat{f}_3(y_{N},\dots,y_{N-d+1})  }$ & ${\scriptstyle  \hat{f}_4(y_{N},\dots,y_{N-d+1}) }$ \\
    \hline
DirRec &  $ {\scriptstyle \hat{f_1}(y_{N},\dots,y_{ N-d+1}) }$ & ${\scriptstyle \hat{f}_2(\hat{y}_{N+1},y_{N},\dots,y_{N-d+1})  }$ & ${\scriptstyle  \hat{f}_3(\hat{y}_{N+2},\hat{y}_{N+1},\dots,y_{N-d+1})  }$ & ${\scriptstyle  \hat{f}_4(\hat{y}_{N+3},\hat{y}_{N+2},\dots,y_{N-d+1})  }$  \\
  \hline
MIMO &  \multicolumn{4}{|c|}{${\scriptstyle  \hat{F}(y_{N},\dots,y_{N-d+1})  }$} \\
  \hline
DIRMO~($s=2$) &  \multicolumn{2}{|c|}{${\scriptstyle  \hat{F}_1(y_{N},\dots,y_{N-d+1})  }$}  & \multicolumn{2}{|c|}{${\scriptstyle  \hat{F}_2(y_{N},\dots,y_{N-d+1})  }$}  \\
  \hline
\end{tabular}
\caption{ \label{exampleStrategies} The different forecasting models used by each strategy to obtain the 4 forecasts needed.}
\end{table}

Let $T_{SO}$ and $T_{MO}$ denote the amount of computational time needed to learn~(with a given learning algorithm) a Single-Output model and a Multiple-Output model, respectively. For a given $H$-step ahead forecasting task, we can see in Table \ref{modelsToLearn} for each strategy the number and type of models to learn, the size of the output for each model as well as the computational time.

\begin{table}[h!]
	\begin{center}
		\begin{tabular}{|c|c|c|c|c|}
			\hline
			 & Number of Models & Types of models & Size of output & Computational time\\ 
			 \hline
			Recursive & 1 & SO & 1 & $1 \times T_{SO}$  \\ 
			\hline
			Direct & $H$ & SO & 1 &  $ H \times T_{SO}$ \\ 
			\hline
			DirRec & $H$ & SO & 1 & $ H \times  (T_{SO}+\mu) $\\ 
			\hline
			MIMO & 1 & MO & $H$ & $1 \times T_{MO}$ \\ 
			\hline
			DIRMO & $\frac{H}{s}$ & MO & $s$ & $\frac{H}{s} \times T_{MO}$\\
			\hline
		\end{tabular}
		\caption{For each forecasting strategy: the number and type of models~(Single-Output or Multiple-Output) to learn  and the size of the output for each model.}
		\label{modelsToLearn}
	\end{center}
\end{table}

Suppose $T_{MO} = T_{SO}  + \delta $, which is a reasonable assumption because learning a model with a vector-valued output takes more time than learning a model with a single-valued output. This allows us to rank the forecasting strategies according to their computation time for training given in Table \ref{modelsToLearn}. Indeed, we have

 \begin{equation} \underbrace{1 \times T_{SO} }_{Recursive} < \underbrace{1 \times T_{MO}}_{MIMO} < \underbrace{\frac{H}{s} \times T_{MO} }_{DIRMO} < \underbrace{H \times T_{SO}}_{Direct} < \underbrace{H \times (T_{SO}+\mu) }_{DirRec} \text{ ,} \label{CtimeStrategies}\end{equation}
 
 \noindent where we suppose that the parameter $s$ of DIRMO is not equal to $1$ or $H$.
 
Note, in one hand, that the time needed to learn a SO model of the DirRec strategy equals $T_{SO}+\mu$ because the input size of each SO task increases at each step. On the other hand, if we need to select the value of the parameter $s$ by on some tuning, the DIRMO strategy will take more time and hence will be the slowest one.
 
In the following, we conclude this section by summarizing the pros and cons of the five forecasting strategies as depicted on Table \ref{summaryStrategies}.
 
\begin{landscape}
\begin{table}
	\begin{center}
		\begin{tabular}{|M{2cm}|M{4cm}|M{4cm}|c|}
			\hline
			 & Pros & Cons & Computational time needed  \tabularnewline
			 \hline
			Recursive & Suitable for noise-free time series (e.g. chaotic)  & Accumulation of errors & $+$  \tabularnewline
			\hline
			Direct & No accumulation of errors & Conditional independence assumption & $++++$  \tabularnewline 
			\hline
			DirRec & Trade-off between Direct and Recursive & Input set grows linearly with $H$ & $+++++$ \tabularnewline 
			\hline
			MIMO & No conditional independence assumption & Reduced flexibility: same model structure for all the horizons & $++$  \tabularnewline
			\hline
			DIRMO & Trade-off between total dependence and total independence of forecasts & One additional parameter to estimate & $+++$ \tabularnewline
			\hline
		\end{tabular}
		\caption{A Summary of the Pros and Cons of the Different Multi-step Forecasting Strategies}
		\label{summaryStrategies}
	\end{center}
\end{table}
\end{landscape}

\clearpage

\section{Lazy Learning for Time Series Forecasting}
\label{sectionLazy}

Each of the forecasting strategies introduced in Section \ref{sectionStrategies} demands the definition of a specific forecasting model or learning algorithm to estimate either the scalar-valued function $f$~(see Equations \ref{stochRec}, \ref{stochDirect} and \ref{stochDirRec}) or the vector-valued function $F$~(see Equations \ref{stochMIMO} and \ref{stochDIRMO}) which represent the temporal stochastic dependencies. As the goal of the paper is not to compare forecasting models~(as in \citep{empiricalMLTimeSeries}) but rather multi-step ahead forecasting strategies, the choice of a underlying forecasting model is required to setup the experiments. In this paper, we adopted the Lazy Learning algorithm, which is a particular instance of local learning models, since it has been shown to be particularly effective in time series forecasting tasks~\citep{Bontempi98BirattariBersini_Leuven,phdbontempi,mimo,mimoijf,mismo,BenTaieb20101950}.

Next section gives a general comparison of global models with local models. Section \ref{LLalgo} presents the Lazy Learning Algorithm in terms of learning properties. Section \ref{SOalgo} and \ref{MOalgo} describe two Lazy Learning algorithms for two types of learning tasks, namely the Single-Output and Multiple-Output Lazy Learning algorithms. Finally, the a discussion is presented on the model combination.

\subsection{Global vs local modeling for supervised learning}

Forecasting the future values of a time series using past observations can be reduced to a supervised learning problem or, more precisely, to a regression problem. Indeed, the time series can be seen as a dataset made of pairs where the first component, called input, is a past temporal pattern and the second, called output, is the corresponding future pattern. Being able to predict the unknown output for a given input is equivalent to forecasting the future values given the last observations of the time series.

\emph{Global modeling} is the typical approach to the supervised learning problem. Global models are parametric models that describe the relationship between the inputs and the output values as an analytical function over the whole input domain. Examples of global models are linear models~\citep{MontgomeryBook2006}, nonlinear statistical regressions~\citep{Seber89Wild} and Neural Networks~\citep{Rumelhart86HintonWilliams}. 

Another approach is the \emph{divide-and-conquer} modeling which consists in relaxing the global modeling assumptions by dividing a complex problem into simpler problems, whose solutions can be combined to yield a solution to the original problem~\citep{phdbontempi}. The divide-and-conquer has evolved in two different paradigms: the \emph{modular architectures} and the \emph{local modeling} approach~\citep{phdbontempi}.

\emph{Modular techniques} replace a global model with different modules covering different parts of the input space. Examples based on this approach are Fuzzy Inference Systems~\citep{Takagi85Sugeno}, Radial Basis Functions~\citep{Moody89Darken,Poggio90Girosi}, Local Model Networks~\citep{MurraySmith94}, Trees~\citep{BreimanFriedmanOlshenStone84} and Mixture of Experts~\citep{Jordan94Jacobs}. The modular approach is in the intermediate scale between the two extremes, the global and the local approach. However, their identification is still performed on the basis of the whole dataset and requires the same procedures used for generic global models.

\emph{Local modeling} techniques are at the extreme end of divide-and-conquer methods. They are nonparametric models that combine excellent theoretical properties with a simple and flexible learning procedure. Indeed, they do not aim to return a complete description of the input/output mapping but rather to approximate the function in a neighborhood of the point to be predicted~(also called the query point). There are different examples of local models, for example nearest neighbor, weighted average, and locally weighted regression~\citep{Atkeson97locallyweighted}. Each of these models use data points near the point to be predicted for estimating the unknown output. \emph{Nearest neighbor} models simply find the closest point and uses its output value. \emph{Weighted average} models combines the closest points by averaging them with weights inversely proportional to their distance to the point to be predicted. \emph{Locally weighted regression} models fit a model to nearby points with a weighted regression where the weights are function of distances to the query point.

The effectiveness of local models is well-known in the time series and computational intelligence community. For example, the method proposed by Sauer~\citep{Sauer94} gave good performance and ranked second best forecast for the Santa Fe A dataset from a forecasting competition organized by Santa Fe institute. Moreover, the two top-ranked entries of the KULeuven competition used local learning methods~\citep{Bontempi98BirattariBersini_Leuven,McNames98}. 

In this work, we will restrict to consider a particular instance of local modeling algorithms: the Lazy Learning algorithm~\citep{Aha97}.

\subsection{The Lazy Learning algorithm}
\label{LLalgo}

It is possible to encounter different degree of ``laziness'' in local learning algorithms. For instance, a $k$ Nearest Neighbor ($k$-NN) algorithm, which learns the best value of $k$ before the query is requested, is hardly a lazy approach since, after the query is presented, it requires only a reduced amount of learning procedure, only the computation of the neighbors and the average. On the contrary, a local method, which depends on the query to select the number of neighbors or other structural parameters presents a higher degree of ``laziness''.

The Lazy Learning~(LL) algorithm, extensively discussed in \citep{Birattari99BontempiBersini,bontempi}, is a \emph{query-based} local modeling technique where \emph{the whole learning procedure} is deferred until a forecast is required. When the query is requested, the learning procedure may start to select the best value of the number of neighbors~(or other structural parameters) and next, the dataset is searched for the nearest neighbors of the query point. The nearest neighbors are then used for estimating a local model, which returns a forecast. The local model is then discarded and the procedure is repeated from scratch for subsequent queries.

The LL algorithm has  a number of attractive features~\citep{Aha97}, namely, the reduced number of assumptions, the online learning capability and the capacity to model nonstationarity. LL assumes no a priori knowledge on the process underlying the data, which is particularly relevant in real datasets. These considerations motivate the adoption of the LL algorithm as a learning model in a multi-step ahead forecasting context.

Local modeling techniques require the definition of a set of model parameters namely \emph{the number $k$ of neighbors}, \emph{the kernel function}, \emph{the parametric family} and \emph{the distance metric}[REF]. In the literature, different methods have been proposed to select automatically the adequate configuration~\citep{Atkeson97locallyweighted,Birattari99BontempiBersini}. However, in this paper, we will limit the search on only the selection of the number of neighbors~(also called or equivalent to the bandwidth selection). This is essentially the most critical parameter, as it controls the bias/variance trade-off. Bandwidth selection is usually performed by rule-of-thumb techniques~\citep{Fan95Gijbels}, plug-in methods~\citep{Ruppert95SheatherWand} or cross-validation strategies~\citep{Atkeson97MooreSchaal}. Concerning the other parameters, we use the tricubic kernel~\citep{Cleveland198887} as kernel function, a constant model for the parametric family and the euclidean distance as metric.

Note that in order to apply local learning to a time series, we need to embed it into a dataset made of pairs where the first component is a temporal pattern of length $d$ and the second component is either the future value~(in the case of Single-Output Modeling) or the consecutive temporal pattern of length $H$~(in the case of Multiple-Output Modeling). In the following sections, $D$ will refer to the embedded time series with $M$ input/output pairs.

\subsection{Single-Output Lazy Learning algorithm}
\label{SOalgo}

In the case of Single-Output learning~(i.e with scalar output), the Lazy Learning procedure consists of a sequence of steps detailed in Algorithm \ref{LLSO}. The algorithm assesses the generalization performance of different local models and compares them in order to select the best one in terms of generalization capability. To perform that, the algorithm associate a Leave-One-Out~(LOO) error $e_{LOO}(k)$ to the estimation $y_{q}^{(k)}$ obtained with $k$ neighbors~(lines \ref{LOOLLSO2} and \ref{LOOLLSO}). 

The LOO error can provide a reliable estimate of the generalization capability. However the disadvantage of such an approach is that it requires to repeat $k$ times the training process, which means a large computational effort. Fortunately, in the case of linear models there exists a powerful statistical procedure to compute the LOO cross-validation measure at a reduced computational cost: the PRESS~(Prediction Sum of Squares) statistic~\citep{Allen1974}.

In case of constant model, the LOO error $e_{LOO}(k)$ for the estimation $y_{q}^{(k)}$ of the query point $\mathbf{x}_{q}$ is calculated as follows~\citep{phdbontempi}:

\begin{equation} e_{LOO}(k)=\frac{1}{k}\sum_{j=1}^{k} (e_{j}(k))^2  \label{pressconst} \text{,}\end{equation}

\noindent where $e_{j}(k)$ designates the error obtained by setting aside the $j$-th neighbor of $\mathbf{x}_{q}$~($j \in \{1,\dots,k\}$). If we define the output of the $k$ closest neighbors of $\mathbf{x}_{q}$ as $\{y_{[1]},\dots,y_{[k]}\}$ then, $e_{j}(k)$ is defined as

\begin{align} 
e_{j}(k) &= y_{[j]} - \frac{\sum_{i=1(i \neq j)}^{k} y_{[i]}}{k-1} \label{looNormal}\\
&= \frac{(k-1)y_{[j]} - \sum_{i=1(i \neq j)}^{k} y_{[i]}}{k-1} \\
&= \frac{(k-1)y_{[j]} +y_{[j]} - y_{[j]} - \sum_{i=1(i \neq j)}^{k} y_{[i]}}{k-1} \\
&= \frac{ky_{[j]} - \sum_{i=1}^{k} y_{[i]}}{k-1} \\
&=  \boxed{ k \frac{y_{[j]}-y_{q}^{(k)}}{k-1} }\text{.} \label{looPRESS}
\end{align}

Note that if we use Equation \ref{looNormal} to calculate the LOO error~(Equation \ref{pressconst}), the training process is repeated $k$ times since the sum in Equation \ref{looNormal} is performed for each index $j$. However, by using the PRESS statistic~(Equation \ref{looPRESS}), we avoid this large computational effort since the sum is replaced by the previously computed $y_{q}^{(k)}$ which was already calculated. This makes the PRESS statistic an efficient method to compute the LOO error.

After evaluating the performance of local models with different number of neighbors $k$~(lines \ref{LLSOFORSTART} to \ref{LLSOFOREND}), the best one which minimizes the LOO error~(having index $k^{*}$) is selected (lines \ref{bestlook} and \ref{bestlook2}). Finally, the prediction of the output of $\mathbf{x}_{q}$ is returned~(line \ref{LOOLLSOreturn}).

\IncMargin{0.1cm}
\begin{algorithm}[h]
\SetKwInOut{Input}{Input}
\SetKwInOut{Output}{Output}
\caption{Single-Output Lazy Learning \label{LLSO}}
\Input{$D=\{(\mathbf{x}_i,y_{i}) \in (\mathbb{R}^{d}\times \mathbb{R}) \text{ with } i \in \{1,\dots,M\}\}$, dataset.}
\Input{$\mathbf{x}_{q} \in \mathbb{R}^{d}$, query point.}
\Input{$Kmax$, the maximum number of neighbors.}
\Output{$\hat{y}_{q}$, the prediction of the (scalar) output of the query point $\mathbf{x}_{q}$.}
\BlankLine
Sort increasingly the set of vectors $\{\mathbf{x}_{i}\}$ with respect to the distance to $\mathbf{x}_{q}$. \\
$[j]$ designate the index of the $j$th closest neighbor of $\mathbf{x}_q$. \\
\BlankLine

 \For{$k \in \{2,\dots,Kmax\}$}{ \nllabel{LLSOFORSTART}
$y_{q}^{(k)}= \frac{1}{k}\sum_{j=1}^{k} y_{[j]}$.\nllabel{LOOLLSO2}  \\
Calculate $e_{LOO}(k)$ which is defined in Equation \ref{pressconst}. \nllabel{LOOLLSO}  \\
} \nllabel{LLSOFOREND}
 $k^{*} = \text{arg min$_{k \in \{2,\dots,Kmax\}}$ } e_{LOO}(k)$. \nllabel{bestlook} \\
 $\hat{y}_{q}=y_{q}^{(k^{*})}$. \nllabel{bestlook2}  \\
\BlankLine
\BlankLine
\Return $\hat{y}_{q}$.\nllabel{LOOLLSOreturn}  \\
\end{algorithm}
\DecMargin{0.1cm}

\subsection{Multiple-Output Lazy Learning algorithm}
\label{MOalgo}

The adoption of Multiple-Output strategies requires the design of multiple-output~(or equivalently multi-response) modeling techniques~\citep{matias,multiresponse,vector-valued} where the output is no more a scalar quantity but a vector of values. Like in the Single-Output case, we need criteria to assess and compare local models with different number of neighbors. In the following, we present two criteria: the first one is  an extension of the LOO error for the Multiple-Output case~(Algorithm \ref{LLMO})~\citep{mimo,BenTaieb20101950} and the second one is a criterion proper to the Multiple-Output modeling~(Algorithm \ref{LLDMO})~\citep{BenTaieb20101950,mimoijf}. Note that, in the two algorithms, the output is a vector of size $l$~(e.g. $l$ will equal $H$ with the MIMO strategy or $s$ in the DIRMO strategy).

\IncMargin{0.1cm}
\begin{algorithm}[h]
\SetKwInOut{Input}{Input}
\SetKwInOut{Output}{Output}

\caption{Multiple-Output Lazy Learning~(\emph{LOO criterion}): MIMO-LOO \label{LLMO}}
\Input{$D=\{(\mathbf{x}_i,\mathbf{y}_{i}) \in (\mathbb{R}^{d}\times \mathbb{R}^{l}) \text{ with } i \in \{1,\dots,M\} \}$, dataset.}
\Input{$\mathbf{x}_{q} \in \mathbb{R}^{d}$, query point.}
\Input{$Kmax$, the maximum number of neighbors.}
\Output{$\mathbf{\hat{y}}_{q}$, the prediction of the (vectorial) output of the query point $\mathbf{x}_{q}$.}
\BlankLine
Sort increasingly the set of vectors $\{\mathbf{x}_{i}\}$ with respect to the distance to $\mathbf{x}_{q}$. \\
$[j]$ will designate the index of the $j$th closest neighbor of $\mathbf{x}_q$. \\
\BlankLine

\For{$k \in \{2,\dots,Kmax\}$}{
$\mathbf{y}_{q}^{(k)}= \frac{1}{k}\sum_{j=1}^{k} \mathbf{y}_{[j]}$. \\
$ E_{LOO}(k)=\frac{1}{l} \sum_{h=1}^{l} e_{LOO}^{l}(k)$ where $e_{LOO}^{l}(k) $ is defined in Equation \ref{pressconst}. \nllabel{LOOLLMO} \\

}
$k^{*} = \text{arg min$_{k \in \{2,\dots,Kmax\}}$ } E_{LOO}(k)$. \\
$\mathbf{\hat{y}}_{q}=\mathbf{y}_{q}^{(k^{*})}$
\BlankLine
\BlankLine
\Return $\mathbf{\hat{y}}_{q}$. \\
\end{algorithm}
\DecMargin{0.1cm}

Algorithm \ref{LLMO} is an extension of the Algorithm \ref{LLSO} for vectorial outputs. We still use the LOO cross-validation measure as a criterion to estimate the generalization capability of the model but here, the LOO error is an aggregation of the errors obtained for each output~(line \ref{LOOLLMO}). Note that the same number of neighbors is selected for all the outputs~(e.g. MIMO strategy) unlike what could happen with different Single-Output tasks~(e.g. Direct strategy).
\clearpage

The second criterion uses the fact that the forecasting horizon $H$ is supposed to be large~(multi-step ahead forecasting) and hence we have enough samples to estimate some descriptive statistics. Then, instead of using the Leave-One-Out error, we can use as criterion a measure of stochastic discrepancy between the forecasted values and the training time series. The lower the discrepancy between the descriptors of the forecasts and the training time series, the better is the quality of the forecasts~\citep{mimoijf}.

Several measures of discrepancy can be defined, both linear and non-linear. For example, the autocorrelation can be used as linear statistics and the maximum likelihood as a non-linear one. In this work, we will consider only one linear measure using both the autocorrelation and the partial correlation. 

The assessement of the quality of the estimation $y_{q}^{(k)}$ of the query point $\mathbf{x}_{q}$ is calculated as follows

\begin{equation} E_{\Delta}(k)= \underbrace{1-|cor[\rho(ts \cdot \mathbf{y}_{q}^{(k)} ),\rho(ts)]| }_{}+ \underbrace{ 1-|cor[\pi(ts \cdot \mathbf{y}_{q}^{(k)}),\pi(ts)]| }_{} \label{dCriterion} \text{,} \end{equation}

\noindent where the symbol ``$\cdot$'' represents the concatenation, $ts$ represent the training time series and $cor$ is the Pearson correlation. This discrepancy measure is composed of two parts where the first part uses the autocorrelation~(noted $\rho$) and the second uses the partial autocorrelation~(noted $\pi$). 

For each part, we calculate the discrepancy~(estimated with the correlation, noted $cor$) between, on one hand, the autocorrelation~(or partial autocorrelation) of the concatenation of the training time series $ts$  and the forecasted sequence $\mathbf{y}_{q}^{(k)}$ and, on the other hand, the autocorrelation~(or partial autocorrelation) of the training time series $ts$~\citep{mimo,mismo}. 

In Algorithm \ref{LLDMO}, after evaluating the performance of local models with different number of neighbors $k$ (lines \ref{LLSOFORSTART}
to \ref{LLSOFOREND}), the best one, which minimizes the discrepancy between the forecasting sequence and the training time series~(having index $k^*$), is selected~(lines \ref{bestd} and \ref{bestd2}). In other words, the goal is to select the best number of neighbors $k^{*}$ which preserve the stochastic properties of the time series in the forecasted sequence. Finally, the prediction of the output of $\mathbf{x}_{q}$ is returned (line \ref{bestpred}).

\IncMargin{0.1cm}
\begin{algorithm}[h]
\SetKwInOut{Input}{Input}
\SetKwInOut{Output}{Output}

\caption{Multiple-Output Lazy Learning~(\emph{discreprancy criterion}): MIMO-ACFLIN \label{LLDMO}}
\Input{$ts=[ts_1,\dots,ts_N]$, time series.}
\Input{$D=\{(\mathbf{x}_i,\mathbf{y}_{i}) \in (\mathbb{R}^{d}\times \mathbb{R}^{l}) \text{ with } i \in \{1,\dots,M\} \}$, dataset.}
\Input{$\mathbf{x}_{q} \in \mathbb{R}^{d}$, query point.}
\Input{$Kmax$, the maximum number of neighbors.}
\Output{$\mathbf{\hat{y}}_{q}$, the prediction of the (vectorial) output of the query point $\mathbf{x}_{q}$.}
\BlankLine
Sort increasingly the set of vectors $\{\mathbf{x}_{i}\}$ with respect to the distance to $\mathbf{x}_{q}$. \\
$[j]$ will designate the index of the $j$th closest neighbor of $\mathbf{x}_q$. \\
\BlankLine

\For{$k \in \{2,\dots,Kmax\}$}{ \label{DFORSTART}
$\mathbf{y}_{q}^{(k)}= \frac{1}{k}\sum_{j=1}^{k} \mathbf{y}_{[j]}$. \\
Calculate $E_{\Delta}(k)$ which is defined in Equation \ref{dCriterion}. \\
} \label{DFOREND}
$k^{*} = \text{arg min$_{k \in \{2,\dots,Kmax\}}$ } E_{\Delta}(k)$. \nllabel{bestd} \\
$\mathbf{\hat{y}}_{q}=\mathbf{y}_{q}^{(k^{*})}$ \nllabel{bestd2}
\BlankLine
\BlankLine
\Return $\mathbf{\hat{y}}_{q}$. \nllabel{bestpred} \\
\end{algorithm}
\DecMargin{0.1cm}

\clearpage

\subsection{Model selection or model averaging}
\label{modelCombination}

Considering the Algorithm \ref{LLSO}, we can see that we generate, for the query $\mathbf{x}_{q}$, a set of predictions $\{y_{q}^{(2)},y_{q}^{(3)},\dots,y_{q}^{(Kmax)}\}$, each obtained with different number of neighbors. For each of these predictions, a testing error $\{e_{LOO}(2),e_{LOO}(3),\dots,e_{LOO}(Kmax)\}$ has been calculated. Note that the next considerations are also applicable to Algorithms \ref{LLMO} and \ref{LLDMO}.

The goal of model selection is to use all this information~(set of predictions and testing errors) to estimate the final prediction $\hat{y}_{q}$ of the query point $\mathbf{x}_{q}$. There exist two main paradigms mainly the \emph{winner-take-all} and \emph{combination} approaches.

In the Algorithm \ref{LLSO}, we presented the winner-take-all approach~(noted WINNER)~\citep{racing97} which consists of comparing the set of models $y_{q}^{(k)}$ and selecting the best one in terms of testing error $e_{LOO}(k)$~(see line \ref{bestlook}).  

Selecting the best model according to the testing error is intuitively the approach which should work the best. However, results in machine learning show that the performance of the final model can be improved by combining models having different structures~\citep{Raudys2006,Jacobs1991,Breiman1996,Schapire1998}.

In order to apply the model averaging, lines $7$ and $8$  of the Algorithm \ref{LLSO} can be replaced by 

\begin{equation}  \hat{y}_{q}=    \frac{ p_{2} \text{ } y_{q}^{(2)} + \dots +  p_{Kmax} \text{ } y_{q}^{(Kmax)}         }  {\sum_{k=2}^{Kmax} p_{k}    }   \text{,} \end{equation}

\noindent where an average is calculated. The weights $p_{k}$ will take different values depending on the combination approach adopted. If $p_{k}$ equals $\frac{1}{Kmax}$, we are in the case of equally weighted combination and $\hat{y}_{q}$ reduces to an arithmetic mean~(noted COMB). Otherwise, if weights are assigned according to testing errors, $p_{k}$ will equal $\frac{1}{e_{LOO}(k)}$ and $\hat{y}_{q}$ reduces to a weighted mean~(noted WCOMB).

\clearpage 

\section{Experimental Setup}
\label{sectionExperiments}

\subsection{Time Series Data}

\begin{figure}[h!]
\centering
\vspace{-0.3cm}
\includegraphics[trim = 0cm 0.5cm 0cm 1.9cm, clip,scale=0.8,scale=0.6]{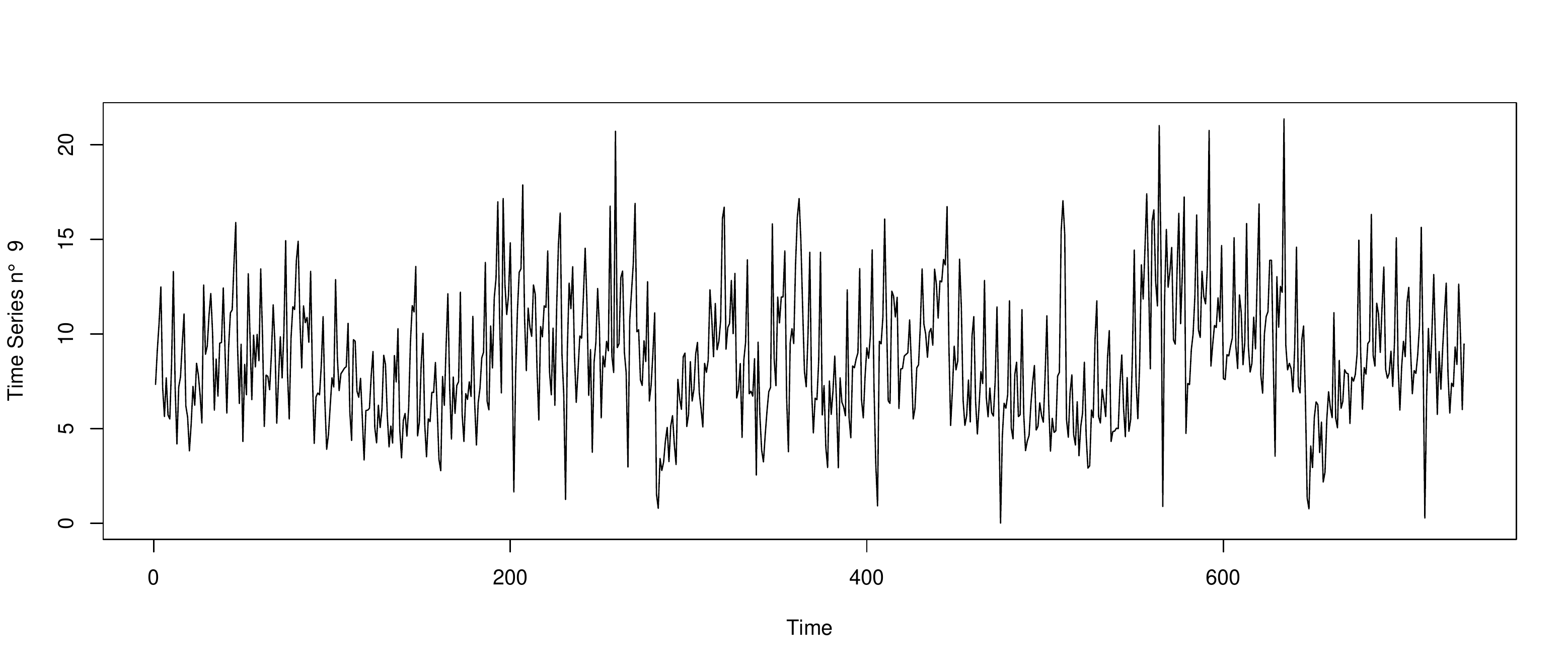} \\
\includegraphics[trim = 0cm 0.5cm 0cm 1.9cm, clip,scale=0.8,scale=0.6]{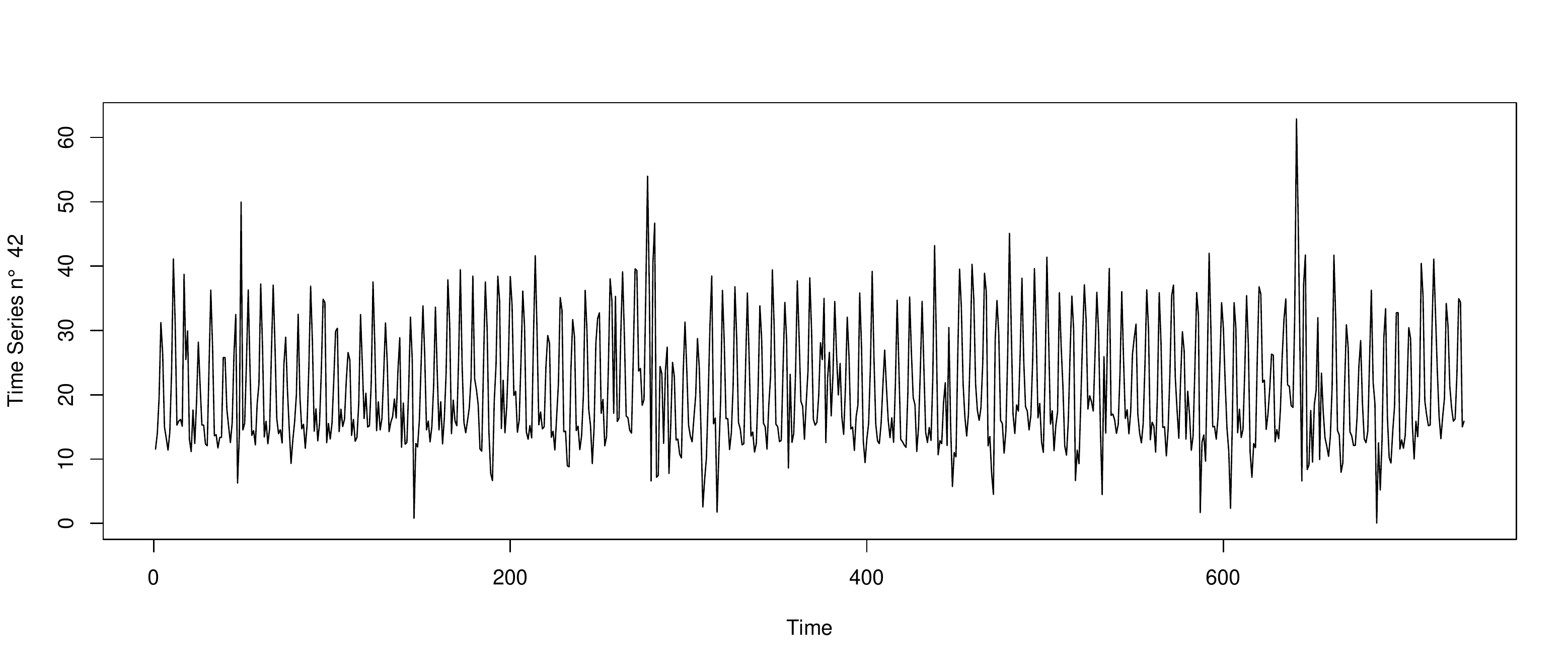} \\
\includegraphics[trim = 0cm 0.5cm 0cm 1.9cm, clip,scale=0.8,scale=0.6]{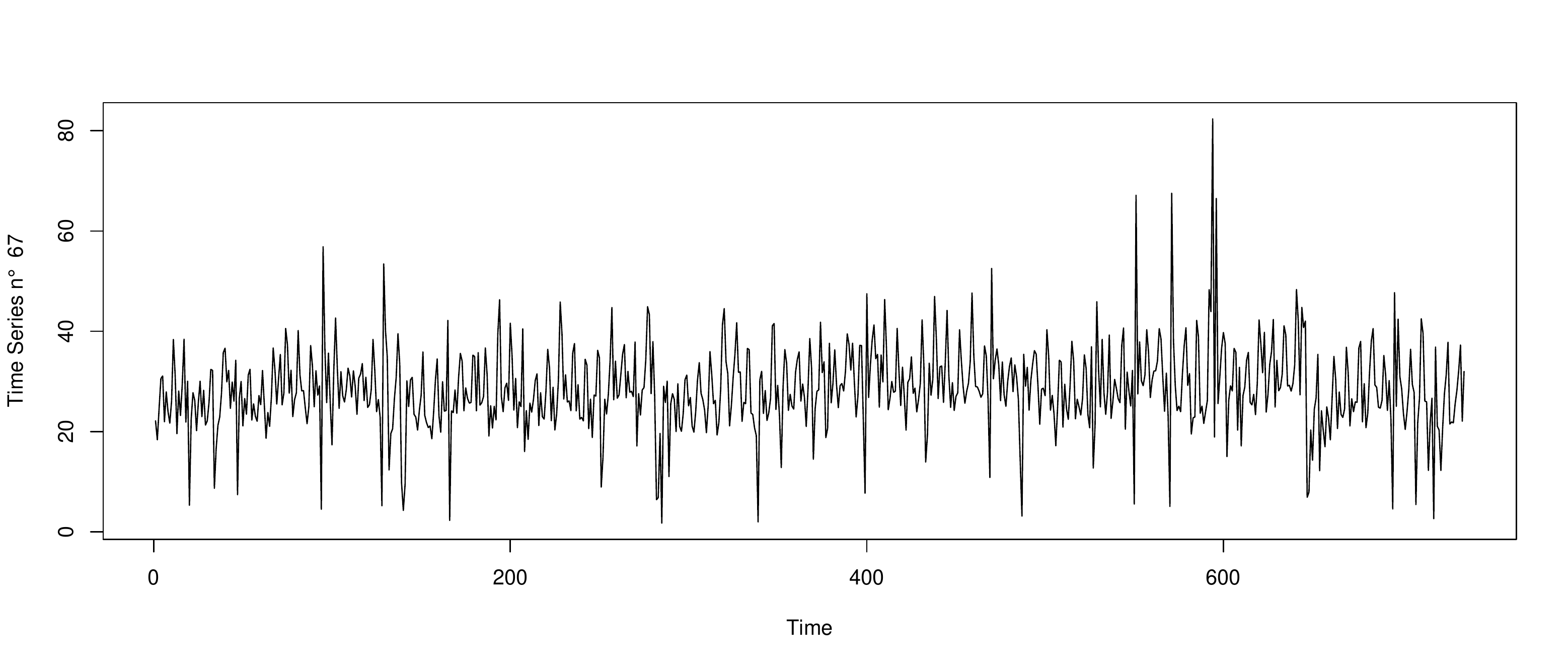} \\
\includegraphics[trim = 0cm 0.5cm 0cm 1.9cm, clip,scale=0.8,scale=0.6]{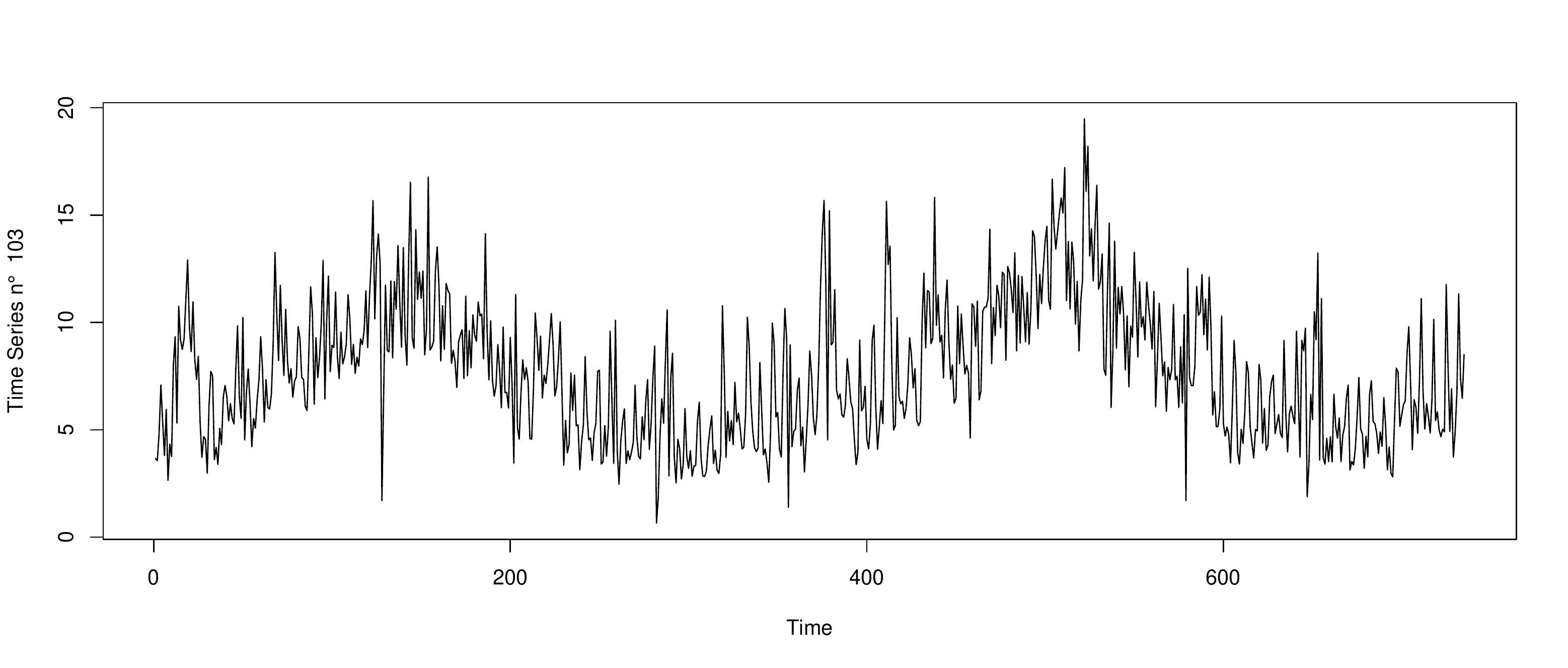} \\
\caption{Four time series from NN5 time series forecasting competition. \label{NN5TimeSeries}}
\end{figure}

In the last decade, several time series forecasting competitions~(e.g. the NN3, NN5, and the ESTSP competitions~\citep{nn3,nn5,estsp07,estsp08}) have been organized in order to compare and evaluate the performance of computational intelligence methods. Among them, the NN5 competition~\citep{nn5} is one of the most interesting one since it includes the challenges of a real-world multi-step ahead forecasting task, namely multiple time series, outliers, missing values as well as multiple overlying seasonalities, etc. Figure \ref{NN5TimeSeries} shows four time series from the NN5 dataset.

Each of the $111$ time series of this competition represents roughly two years of daily cash money withdrawal amounts~($735$ data points) at ATM machines at one of the various cities in the UK. For each time series, the competition required to forecast the values of the next $56$ days, using the given historical data points, as accurately as possible. The performance of the forecasting methods over one time series was assessed by the symmetric mean absolute percentage of error~(SMAPE) measure~\citep{nn5}, defined as

\begin{equation} \text{SMAPE} = \frac{1}{H} \sum_{h=1}^{H} \frac{|\hat{y}_h-y_h|}{(\hat{y}_h+y_h)/2} \times 100 \textnormal{,} \end{equation}

\noindent where $y_h$ is the target output and $\hat{y}_h$ is the prediction. Since this is a relative error measure, the errors can be averaged over all time series to obtain a mean SMAPE defined as

\begin{equation} \text{SMAPE}^{*} = \frac{1}{111} \sum_{i=1}^{111} \text{SMAPE}_{i} \textnormal{,} \label{smapeCriterion} \end{equation}

\noindent where SMAPE$_{i}$ denotes the SMAPE of the $i$th time series. 

\clearpage

\clearpage

\subsection{Methodology}
\label{secMethodo}

\begin{figure}[h!]
\centering
\includegraphics[scale=0.4]{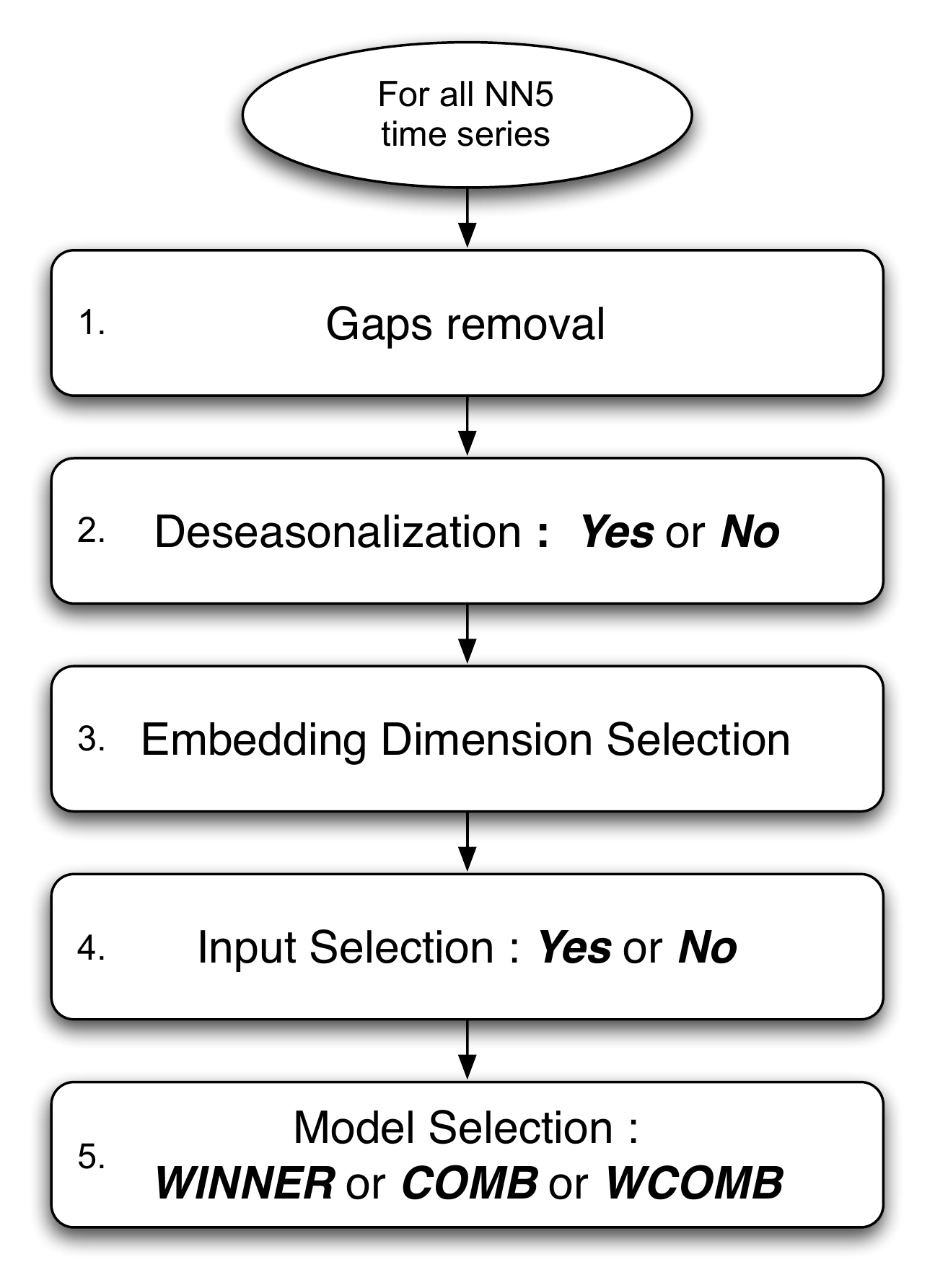} 
\caption{The Different Preprocessing Steps. \label{methodoGeneral}}
\end{figure}

The aim of the experimental study is to compare the accuracy of the five forecasting strategies in the
context of the NN5 competition. Since the accuracy of a forecasting technique is known to be dependent on several design choices (e.g. the deseasonalization or the input selection) and we want to focus our analysis on the multi-step ahead forecasting strategies, we consider a number of different configurations in order to increase the statistical power of our comparison. Every configuration is composed of several preprocessing steps as sketched in Figure  \ref{methodoGeneral}. Since some of  these steps (e.g. deseasonalization) can be performed in alternative ways (e.g. two alternatives for the deseasonalization, two alternatives for input selection, three alternatives for  the model selection), we come up with $12$ configurations. The details about each step are given in what follows.

\noindent \textbf{Step 1: Gaps removal} 

The specificity of the NN5 series requires a preprocessing step called \emph{gaps removal} where by gap we mean two types of 
anomalies:  (i) zero values that indicate that no money withdrawal occurred and (ii) missing observations for which no value was recorded. 
About  $2.5 \%$ of the data are corrupted by gaps. In our experiments we adopted the gap removal method proposed in~\citep{Wichard2010}: if $y_m$ is the gap sample, this method replaces the gap with the median of the set $[y_{m+365}, y_{m-365}, y_{m+7} \text{ and } y_{m-7}]$ among which are available.

\noindent \textbf{Step 2: Deseasonalization}
The adoption of deseasonalization may have a large impact on the forecasting strategies
because the NN5 time series possess a variety of periodic patterns. 
For that reason we decided to consider tasks with and without deseasonalization in order to better account for
the role of the forecasting strategy. 
We adopt the deseasonalization methodology discussed in \citep{amirnn5} to remove the strong day of the week seasonality as well as the moderate day of the month seasonality. Of course after we deseasonalize and apply the forecasting model
we restore back the seasonality.  

\noindent \textbf{Step 3: Embedding dimension selection}

Every forecasting strategy requires the setting of the size $d$  of the embedding dimension~(see Equations \ref{stochRec} to \ref{stochDIRMO}). Several approaches have been proposed in the literature to select this value~\citep{nonlinear}. Since this aspect is not a central theme 
in our paper we just applied the state-of-the-art approach reviewed in \citep{Crone:2009:ISN:1704555.1704739}, which consists of selecting the  time-lagged realizations with significant partial correlation function~(PACF). This method allows to select the value of the embedding dimension and then to identify the relevant variables within the window of past observations. We set the maximum lag of the PACF to $200$ to provide a sufficiently comprehensive pool of features. However, note that the final dimensionality of the input vectors for all the time series is on average equal to $24$.

\noindent \textbf{Step 4: Input Selection}

We considered the forecasting task with and without input variable selection step.
A variable selection procedure requires the setting of  two elements: the \emph{relevance criterion}, i.e. statistics which estimates the quality of the selected variables, and the \emph{search procedure}, which describes the policy to explore the input space.
We adopted the \emph{Delta test}(DT) as  relevance criterion. The DT has been introduced in time series forecasting domain by Pi and Peterson in \citep{Pi94} and later successfully applied to several forecasting task~\citep{mismo,BenTaieb20101950,Liitiainen07,guillen08newdelta,mateo10_neuro}. This criterion is based on applying some kind of a noise variance estimator, and then selecting the set of input variables that yield the strongest and most deterministic dependence between inputs and outputs~\citep{Mateo08}.





Concerning the search procedure, we adopted a Forward-Backward Search (FBS) procedure which is a combination of forward selection (sequentially adding input variables) and backward search (sequentially removing some input variables). This choice was motivated by the flexibility of the FBS procedure which allows a deeper exploration of the input space. Note that the search is initialized by the set of variables
defined in the previous step.

\noindent \textbf{Step 5: Model Selection}

Concerning the model selection procedure, three approaches~(see Section \ref{modelCombination}) are taken into consideration in our experiments:

\begin{description}
\item [WINNER]: This approach selects the model that gives best performance for the test set (winner-take-all approach). 
\item [COMB]: This approach combines all alternative models by simple averaging. 
\item [WCOMB]: This approach combines models by weighted averaging where weights are inversely proportional to the test errors. 

\end{description}

\subsubsection{The Compared forecasting strategies}

Table \ref{tableStrategies} presents the eight forecasting strategies that we tested, showing also their respective acronyms.
\begin{table}[h]
\centering
\begin{tabular}{|l|M{4cm}|M{6cm}|} 
\hline
\cline{1-3}
$1.$ REC  & \multicolumn{2}{l|}{The \emph{Recursive} forecasting strategy.}  \tabularnewline
\hline
\cline{1-3}
$2.$ DIR & \multicolumn{2}{l|}{The \emph{Direct} forecasting strategy.}  \tabularnewline
\hline
\cline{1-3}
$3.$ DIRREC & \multicolumn{2}{l|}{The \emph{DirRec} forecasting strategy.}  \tabularnewline
\hline
\cline{1-3}
\multicolumn{1}{|c|}{ \multirow{2}{*}{\vspace{-1cm} MIMO} }  & $4.$ MIMO-LOO & A variant of the \emph{MIMO} forecasting strategy with the LOO selection criteria. \tabularnewline
\cline{2-3}
  & $5.$ MIMO-ACFLIN & A variant of the \emph{MIMO} forecasting strategy with the autocorrelation selection criteria. \tabularnewline
\hline 
\cline{1-3}
\multicolumn{1}{|c|}{ \multirow{3}{*}{\vspace{-2.4cm} DIRMO } }  & $6.$ DIRMO-SEL & The \emph{DIRMO} forecasting strategy which select the best value of the parameter $s$. \tabularnewline
  \cline{2-3}
  & $7.$ DIRMO-AVG & A variant of the \emph{DIRMO} strategy which calculates a simple average of the forecasts obtained with different values of the parameter $s$. \tabularnewline
  \cline{2-3}
  & $8.$ DIRMO-WAVG & The DIRMO-AVG with a weighted average where weights are inversely proportional to testing errors. \tabularnewline
\hline
\cline{1-3}
\end{tabular}
\caption{The five forecasting strategies with their respective variants which gives eight forecasting strategies. \label{tableStrategies}}
\end{table}

\clearpage

\subsubsection{Forecasting performance evaluation}
\label{sectionFevaluation}
\begin{figure}[h!]
\centering
\includegraphics[scale=0.5]{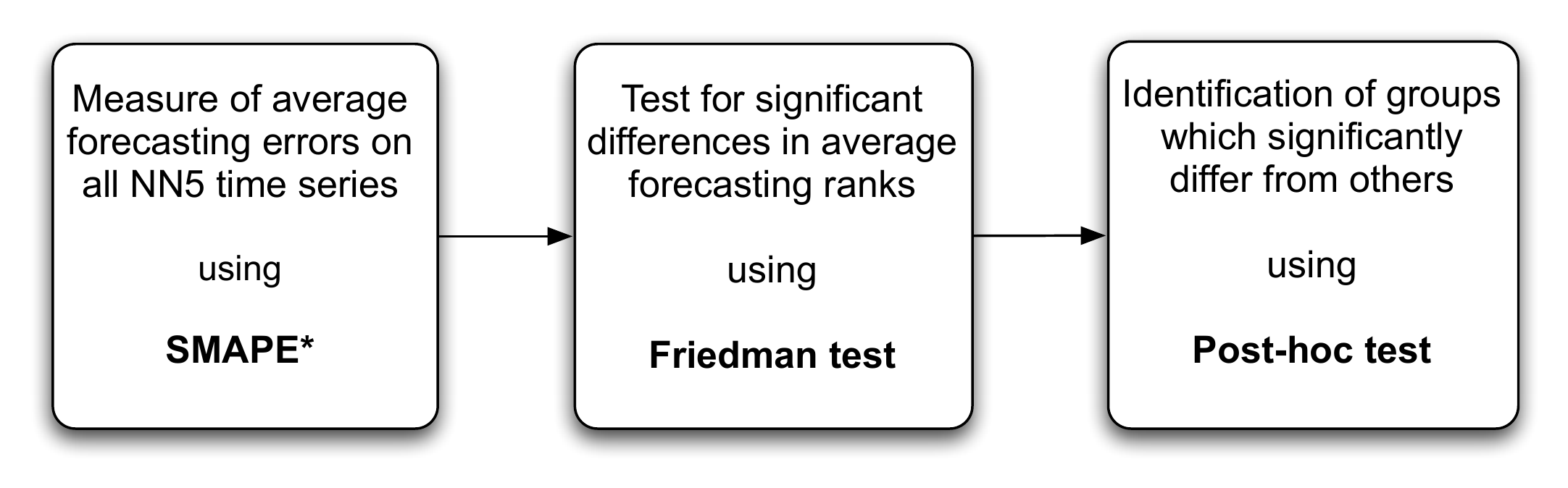} 
\caption{Different steps of the forecasting performance evaluation. \label{PerformanceEvaluation}}
\end{figure}

This section describes the assessment procedure (Figure~\ref{PerformanceEvaluation}) of the 8 forecasting strategies.  

The procedure for comparing between the eight forecasting strategies is shown in Figure~\ref{PerformanceEvaluation}. The accuracy of each forecasting strategy is first measured using the SMAPE* measure calculated over the $111$ time series and defined in Equation \ref{smapeCriterion}. 
To test if there are significant general differences
in performance between the different strategies,
we have to consider the problem of comparing multiple models
on multiple data sets. For such case Dem\v{s}ar ~\citep{Demsar_statisticalcomparisons,extensionDemsar} in a detailed comparative study 
recommended using a two stage procedure: first to apply 
Friedman's or Iman and Davenport's tests to test if the compared models
have the same mean rank. If this test rejects the null-hypothesis, then 
post-hoc pairwise tests are to be performed to compare
the different models. These tests adjust the critical values higher
to ensure that there is at most a 5\% chance that one of
the pairwise differences will be erroneously found significant.

\noindent \textbf{Friedman test}

The Friedman test~\citep{friedman1937,friedman1940} is a non-parametric procedure which tests the significance of differences between multiple ranks. It ranks the algorithms for each dataset separately: the rank of 1 will be given to the best performing algorithm, the rank of 2 to the second best and so on. Note that average ranks are assigned in case of ties.

After ranking the algorithms for each dataset, the Friedman test compares the average ranks of algorithms. Let $r^{i}_{j}$ be the rank of the $j$-th of $k$ algorithms on the $i$-th of $N$ data sets, the average rank of the $j$-th algorithm is $R_{j}=\frac{1}{N}\sum_{i} r^{i}_{j}$.

The null-hypothesis states that all the algorithms are equivalent and so their ranks $R_{j}$ should be equal. Under the null-hypothesis, the Friedman statistic 

\begin{equation}
Q = \frac{12N}{k(k+1)} \left[  \sum_{j} R_{j}^{2} - \frac{k(k+1)^2 }{4}  \right]
\end{equation}

\noindent is distributed according to a chi-squared with $k-1$ degrees of freedom~($\chi^{2}_{k-1}$), when $N$ and $k$ are large enough~(as a rule of a thumb, $N>10$ and $k>5$)~\citep{Demsar_statisticalcomparisons}.

Iman and Danvenport~\citep{Iman80approximationsof}, showing that Friedman's statistic is undesirably conservative, derived another
improved statistic, given by

\begin{equation}
S = \frac{(N-1) Q}{N(k-1)-Q} 
\end{equation}

\noindent which is distributed, under the null-hypothesis, according to the F-distribution with $k-1$ and $(k-1)(N-1)$ degrees of freedom.

\noindent \textbf{Post-hoc test}

When the null-hypothesis is rejected, i.e. there is a significant difference between at least 2 strategies, a post-hoc test is performed to identify significant pairwise differences among all the algorithms. The test statistic for comparing the $i$-th and the $j$-th algorithm is 

\begin{equation}
z=\frac{(R_{i}-R_{j})}{\sqrt{\frac{k(k+1)}{6N}}} \text{,}
\end{equation}

\noindent which is asymptotically normally distributed under the null hypothesis. After the corresponding $p$-value is calculated, it is compared with a given level of significance $\alpha$.

However, in multiple comparisons, as there are a possibly large number of pairwise comparisons, there is a relatively high chance that some pairwise test are incorrectly rejected. Several procedures exist to adjust the value of $\alpha$ to compensate for this bias, for instance Nemenyi, Holm, Shaffer as well as Bergmann and Hommel~\citep{Demsar_statisticalcomparisons}. Based on the suggestion of Garcia and Herrera~\citep{extensionDemsar} we adopted Shaffer's correction. The reason is that Garcia and Herrera~\citep{extensionDemsar} showed that Shaffer's procedure has the same complexity Holm's procedure, but with the advantage of using information about logically related hypothesis.

\subsection{Experimental phases}

In order to reproduce the same context of the NN5 forecasting competition the experimental setting is made of two phases: the \emph{pre-competition} and the \emph{competition} phase.

\subsubsection{Pre-competition phase}

The pre-competition phase is devoted to the comparison of the different forecasting strategies using the available observations of $111$ time series. The goal is to learn the different parameters and then estimate the forecasting performance and compare between the different strategies. 

\begin{figure}[h!]
\centering
\includegraphics[trim = 0cm 0.5cm 0cm 1.9cm, clip,scale=0.8,scale=0.6]{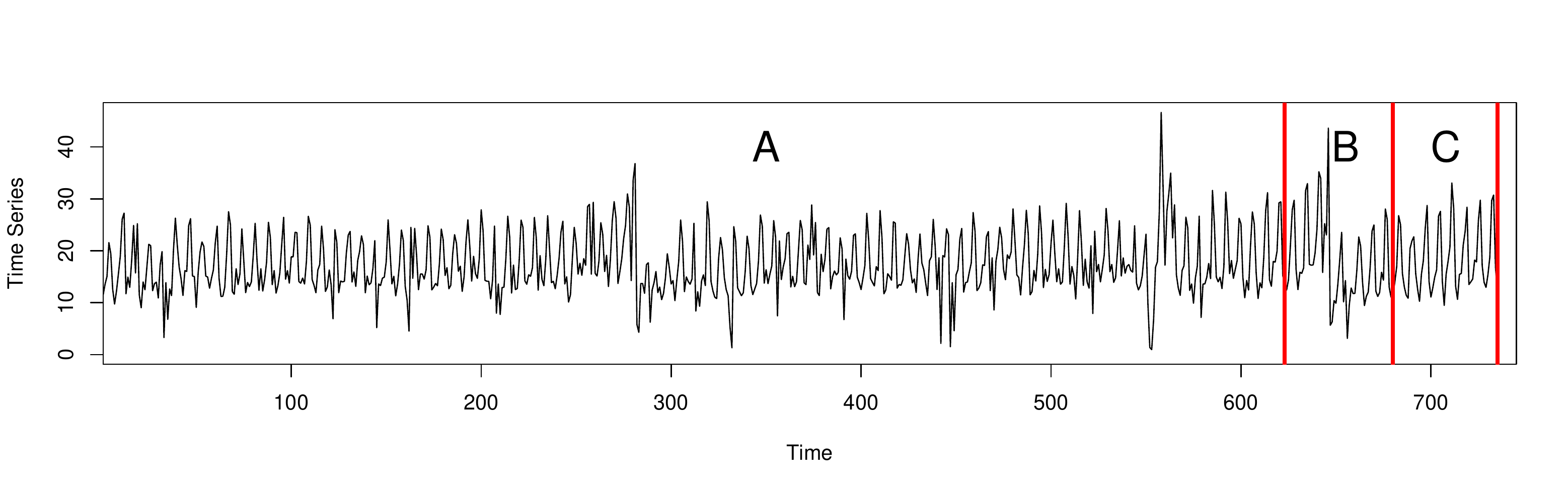} 
\caption{Learning with three mutually exclusive sets for training~(A), validation~(B) and testing~(C).\label{lPeriod}}
\end{figure}

To estimate the forecasting performance of each strategy, we used a learning scheme with training-validation-testing sets. Each time series~(containing $735$ observations) is partitioned in three mutually exclusive sets~(A, B and C) as shown in Figure \ref{lPeriod}: training~(Day $1$ to Day $623$: $623$ values), validation~(Day $624$ to Day $679$: $56$ values) and testing~(Day $680$ to Day $735$: $56$ values). 

The validation set~(B in Figure \ref{lPeriod}) is used to build and tune the models. Specifically, as we use a Lazy Learning approach, we need  to select, for each model, the range of number of neighbors($[2,\dots,Kmax]$) to use in performing the forecasting task. 

The test set~(C in Figure \ref{lPeriod}) is used to measure the performances of each forecasting strategy. To make the utmost use of the available data, we adopt a multiple time origin test as suggested by Tashman in~\citep{Tashman00}, where the time origin denotes the point from which the multi-step ahead forecasts are generated.

The time origin and corresponding forecast intervals are given as:

\begin{enumerate}
\item Day $680$ to Day $735$~($56$ data points)
\item Day $687$ to Day $735$~($49$ data points)
\item Day $694$ to Day $735$~($42$ data points)
\end{enumerate}

In other words, we perform the forecast three times starting from the three different starting points, each time forecasting
a number of steps ahead till the end of the interval. 
Note that we used the same test period and evaluation criterion (i.e. the SMAPE) as used by Andrawis et al in~\citep{amirnn5}. This allows us to compare our results with several other machine learning models tested in this article.

\subsubsection{Competition phase}

\begin{figure}[h!]
\centering
\includegraphics[trim = 0cm 0.5cm 0cm 1.9cm, clip,scale=0.8,scale=0.6]{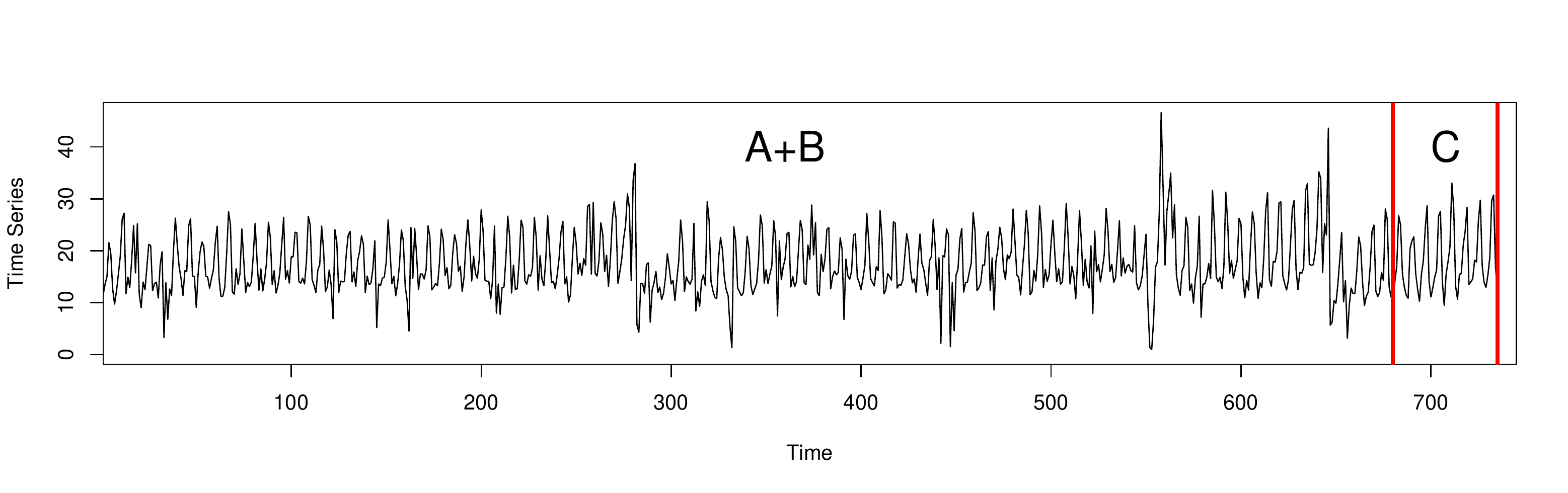} 
\caption{Forecasting with two mutually exclusive sets for training~(A+B) and validation~(C). \label{fPeriod}}
\end{figure}

In the competition phase we generate the final forecasts, made up of $56$ future observations, which would have been submitted to the competition. This phase takes advantage of the lessons learned and the design choices made in the pre-competition phase. Here, we combine the training set with the test set~(A+B in Figure \ref{fPeriod}) to retrain the models of the different strategies and then generate the final forecast~(which will be submitted to the competition). The training set~(A+B on Figure \ref{fPeriod}) is now made of $679$ data points and the validation set~(C on Figure \ref{fPeriod}) is composed of the next $56$ data points, as shown in Figure \ref{fPeriod}. In other words, the $735$ values are then used to build and tune the models, which will next return the forecasted values.

\clearpage

\section{Results and discussion}
\label{sectionResDiscussion}

This section presents and discusses the prediction results of the forecasting strategies for the \emph{pre-competition} and \emph{competition} phase. For each phase, we report the results obtained in the $12$ different configurations introduced in Section~\ref{secMethodo}.

The forecasting performance of the strategies is measured using the criteria discussed in Section~\ref{sectionFevaluation} and presented by means of two tables.  The first one provides the average SMAPE as well as the ranking for each forecasting strategy. Since the null-hypothesis stating that all the algorithms are equivalent
has been rejected (using the Iman-Davenport statistic) for all the configurations, we proceeded with the post-hoc test. The second table presents the results of this post-hoc test, which partitions the set of strategies in several groups which are statistically signicantly different in terms of forecasting performance.

Note that the configurations which require the input selection do not contain the DIRMO results since combining the selection of the inputs with the selection of the parameter $s$ would have needed an excessive amount of computation time.

\subsection{Pre-competition results}

The SMAPE and ranking results for the \emph{pre-competition} phase are presented in Table~\ref{tableErrors.precompetition} while the results of the post-hoc test are summarized in Table~\ref{posthoc.precompetition}.

%
\begin{table}[h!]
\setlength{\extrarowheight}{3pt}
\scriptsize
\begin{center}
\subfloat[]{
\label{tableErrors.precompetition.WithSeas.NoInputSel}
\begin{tabular}{|l||c|c||c|c||c|c|} 
\cline{2-7}
\multicolumn{1}{c}{} &\multicolumn{6}{|c|}{\large Deseasonalization : {\bf No} - Input Selection : {\bf No} }\\ 
\hline 
\multicolumn{1}{|l|}{\bf Strategies}&\multicolumn{2}{c||}{\bf WINNER}&\multicolumn{2}{c||}{\bf COMB}&\multicolumn{2}{c|}{\bf COMBW}\\ \cline{2-7}
\multicolumn{1}{|l|}{}&\multicolumn{1}{c|}{SMAPE* [Std err]}&\multicolumn{1}{c||}{Avg Rank}&\multicolumn{1}{c|}{SMAPE* [Std err]}&\multicolumn{1}{c||}{Avg Rank}&\multicolumn{1}{c|}{SMAPE* [Std err]}&\multicolumn{1}{c|}{Avg Rank}\\ \hline
DIR&22.37[0.55](7)&5.19(7)&22.19[0.54](7)&4.96(6)&22.61[0.55](5)&5.47(7)\\ 
REC&21.41[0.59](6)&3.98(5)&21.95[0.58](5)&4.63(5)&22.91[0.62](7)&4.73(5)\\ 
DIRREC&45.25[0.89](8)&8.00(8)&40.73[0.83](8)&8.00(8)&38.93[0.77](8)&7.99(8)\\ 
\hline
MIMO-LOO&21.17[0.60](2)&3.64(2)&{\bf20.61}[0.57](1)&{\bf2.77}(1)&{\bf20.61}[0.58](1)&{\bf2.71}(1)\\ 
MIMO-ACFLIN&21.40[0.59](5)&4.48(6)&20.69[0.58](2)&3.17(2)&20.61[0.58](2)&2.71(2)\\ 
DIRMO-SEL&21.21[0.56](3)&3.76(3)&22.15[0.56](6)&5.40(7)&22.68[0.56](6)&5.32(6)\\ 
DIRMO-AVG&21.27[0.56](4)&3.88(4)&20.90[0.56](3)&3.36(3)&21.16[0.56](3)&3.42(3)\\ 
DIRMO-WAVG&{\bf21.12}[0.56](1)&{\bf3.07}(1)&20.96[0.56](4)&3.71(4)&21.25[0.57](4)&3.65(4)\\ 
\hline
\end{tabular}

}\\
\subfloat[]{
\label{tableErrors.precompetition.WithSeas.InputSel}
\begin{tabular}{|l||c|c||c|c||c|c|} 
\cline{2-7}
\multicolumn{1}{c}{} &\multicolumn{6}{|c|}{\large Deseasonalization : {\bf No} - Input Selection : {\bf Yes} }\\ 
\hline 
\multicolumn{1}{|l|}{\bf Strategies}&\multicolumn{2}{c||}{\bf WINNER}&\multicolumn{2}{c||}{\bf COMB}&\multicolumn{2}{c|}{\bf COMBW}\\ \cline{2-7}
\multicolumn{1}{|l|}{}&\multicolumn{1}{c|}{SMAPE* [Std err]}&\multicolumn{1}{c||}{Avg Rank}&\multicolumn{1}{c|}{SMAPE* [Std err]}&\multicolumn{1}{c||}{Avg Rank}&\multicolumn{1}{c|}{SMAPE* [Std err]}&\multicolumn{1}{c|}{Avg Rank}\\ \hline
DIR&22.80[0.50](4)&3.20(4)&22.73[0.52](4)&3.27(4)&23.37[0.54](4)&3.27(4)\\ 
REC&21.17[0.55](2)&2.20(2)&22.21[0.52](3)&2.96(3)&23.20[0.56](3)&3.01(3)\\ 
DIRREC&45.21[0.93](5)&5.00(5)&40.57[0.83](5)&4.96(5)&39.03[0.75](5)&4.96(5)\\ 
\hline
MIMO-LOO&{\bf21.06}[0.57](1)&{\bf2.02}(1)&{\bf20.60}[0.59](1)&{\bf1.79}(1)&{\bf20.64}[0.59](1)&{\bf1.88}(1)\\ 
MIMO-ACFLIN&21.40[0.57](3)&2.58(3)&20.65[0.59](2)&2.01(2)&20.64[0.59](2)&1.88(2)\\ 
\hline
\end{tabular}

}\\
\subfloat[]{
\label{tableErrors.precompetition.WithoutSeas.NoInputSel}
\begin{tabular}{|l||c|c||c|c||c|c|} 
\cline{2-7}
\multicolumn{1}{c}{} &\multicolumn{6}{|c|}{\large Deseasonalization : {\bf Yes} - Input Selection : {\bf No} }\\ 
\hline 
\multicolumn{1}{|l|}{\bf Strategies}&\multicolumn{2}{c||}{\bf WINNER}&\multicolumn{2}{c||}{\bf COMB}&\multicolumn{2}{c|}{\bf COMBW}\\ \cline{2-7}
\multicolumn{1}{|l|}{}&\multicolumn{1}{c|}{SMAPE* [Std err]}&\multicolumn{1}{c||}{Avg Rank}&\multicolumn{1}{c|}{SMAPE* [Std err]}&\multicolumn{1}{c||}{Avg Rank}&\multicolumn{1}{c|}{SMAPE* [Std err]}&\multicolumn{1}{c|}{Avg Rank}\\ \hline
DIR&22.29[0.57](7)&6.42(8)&20.41[0.57](7)&6.23(8)&20.49[0.58](7)&6.43(8)\\ 
REC&21.61[0.61](6)&5.35(6)&19.39[0.59](6)&3.87(4)&19.31[0.59](5)&3.68(2)\\ 
DIRREC&22.61[0.68](8)&6.33(7)&20.67[0.64](8)&6.15(7)&20.61[0.61](8)&6.04(7)\\ 
\hline
MIMO-LOO&19.81[0.59](4)&3.83(4)&19.21[0.60](4)&{\bf3.69}(1)&19.25[0.60](3)&3.82(4)\\ 
MIMO-ACFLIN&{\bf19.34}[0.59](1)&3.21(3)&19.19[0.59](3)&4.00(5)&19.25[0.60](4)&3.82(5)\\ 
DIRMO-SEL&19.49[0.57](2)&{\bf2.90}(1)&{\bf18.98}[0.57](1)&3.72(2)&{\bf19.04}[0.58](1)&{\bf3.66}(1)\\ 
DIRMO-AVG&20.39[0.56](5)&4.74(5)&19.36[0.58](5)&4.61(6)&19.43[0.58](6)&4.77(6)\\ 
DIRMO-WAVG&19.71[0.58](3)&3.21(2)&19.02[0.57](2)&3.72(3)&19.11[0.58](2)&3.79(3)\\ 
\hline
\end{tabular}

}\\
\subfloat[]{
\label{tableErrors.precompetition.WithoutSeas.InputSel}
\begin{tabular}{|l||c|c||c|c||c|c|} 
\cline{2-7}
\multicolumn{1}{c}{} &\multicolumn{6}{|c|}{\large Deseasonalization : {\bf Yes} - Input Selection : {\bf Yes} }\\ 
\hline 
\multicolumn{1}{|l|}{\bf Strategies}&\multicolumn{2}{c||}{\bf WINNER}&\multicolumn{2}{c||}{\bf COMB}&\multicolumn{2}{c|}{\bf COMBW}\\ \cline{2-7}
\multicolumn{1}{|l|}{}&\multicolumn{1}{c|}{SMAPE* [Std err]}&\multicolumn{1}{c||}{Avg Rank}&\multicolumn{1}{c|}{SMAPE* [Std err]}&\multicolumn{1}{c||}{Avg Rank}&\multicolumn{1}{c|}{SMAPE* [Std err]}&\multicolumn{1}{c|}{Avg Rank}\\ \hline
DIR&21.87[0.47](4)&3.95(5)&20.07[0.49](4)&3.82(5)&20.20[0.51](4)&3.97(5)\\ 
REC&20.52[0.53](3)&2.95(3)&19.14[0.55](3)&2.57(3)&19.20[0.55](3)&2.61(3)\\ 
DIRREC&23.06[0.84](5)&3.77(4)&21.03[0.68](5)&3.77(4)&20.95[0.66](5)&3.74(4)\\ 
\hline
MIMO-LOO&20.02[0.53](2)&2.59(2)&18.86[0.54](2)&2.45(2)&{\bf18.88}[0.54](1)&{\bf2.34}(1)\\ 
MIMO-ACFLIN&{\bf18.95}[0.54](1)&{\bf1.76}(1)&{\bf18.81}[0.54](1)&{\bf2.38}(1)&18.88[0.54](2)&2.34(2)\\ 
\hline
\end{tabular}

}
\caption{\emph{\textbf{Pre-competition phase}}. Average forecasting errors (SMAPE*) with average ranking for each strategy in the 12 different configurations. The numbers in round bracket represent the ranking within the column. \label{tableErrors.precompetition}}
\end{center}
\end{table}

\begin{table}[!htp]
\centering\tiny
\subfloat[]{
\label{posthoc.precompetition.WithSeas.NoInputSel}
\begin{tabular}{|c|c|c|}
\hline
\multicolumn{3}{|c|}{\begin{minipage}{0.6\textwidth}\medskip \large Deseasonalization : {\bf No} - Input Selection : {\bf No} \end{minipage}} \\\hline
\textbf{WINNER} & \textbf{COMB} & \textbf{WCOMB} \\\hline

\begin{minipage}{.2\textwidth}
\begin{enumerate}[leftmargin=1em]
\item "DIRMO-WAVG" (3.07) \\"MIMO-LOO" (3.64) \\"DIRMO-SEL" (3.76) \\"DIRMO-AVG" (3.88) \\"REC" (3.98) \\"MIMO-ACFLIN" (4.48) \\"DIR" (5.19) \\
\item "DIRREC" (8.00) \\
\end{enumerate}
\end{minipage} 
 &
\begin{minipage}{.2\textwidth}
\begin{enumerate}[leftmargin=1em]
\item "MIMO-LOO" (2.77) \\"MIMO-ACFLIN" (3.17) \\"DIRMO-AVG" (3.36) \\"DIRMO-WAVG" (3.71) \\
\item "REC" (4.63) \\"DIR" (4.96) \\"DIRMO-SEL" (5.40) \\
\item "DIRREC" (8.00) \\
\end{enumerate}
\end{minipage} 
 &
\begin{minipage}{.2\textwidth}
\begin{enumerate}[leftmargin=1em]
\item "MIMO-LOO" (2.71) \\"MIMO-ACFLIN" (2.71) \\"DIRMO-AVG" (3.42) \\"DIRMO-WAVG" (3.65) \\
\item "REC" (4.73) \\"DIRMO-SEL" (5.32) \\"DIR" (5.47) \\
\item "DIRREC" (7.99) \\
\end{enumerate}
\end{minipage}\\ 
 
\hline 
\end{tabular}

}\\
\subfloat[]{
\label{posthoc.precompetition.WithSeas.InputSel}
\begin{tabular}{|c|c|c|}
\hline
\multicolumn{3}{|c|}{\begin{minipage}{0.6\textwidth}\medskip \large Deseasonalization : {\bf No} - Input Selection : {\bf Yes} \end{minipage}} \\\hline
\textbf{WINNER} & \textbf{COMB} & \textbf{WCOMB} \\\hline

\begin{minipage}{.2\textwidth}
\begin{enumerate}[leftmargin=1em]
\item "MIMO-LOO" (2.02) \\"REC" (2.20) \\"MIMO-ACFLIN" (2.58) \\
\item "DIR" (3.20) \\
\item "DIRREC" (5.00) \\
\end{enumerate}
\end{minipage} 
 &
\begin{minipage}{.2\textwidth}
\begin{enumerate}[leftmargin=1em]
\item "MIMO-LOO" (1.79) \\"MIMO-ACFLIN" (2.01) \\
\item "REC" (2.96) \\"DIR" (3.27) \\
\item "DIRREC" (4.96) \\
\end{enumerate}
\end{minipage} 
 &
\begin{minipage}{.2\textwidth}
\begin{enumerate}[leftmargin=1em]
\item "MIMO-LOO" (1.88) \\"MIMO-ACFLIN" (1.88) \\
\item "REC" (3.01) \\"DIR" (3.27) \\
\item "DIRREC" (4.96) \\
\end{enumerate}
\end{minipage}\\ 
 
\hline 
\end{tabular}

}\\
\subfloat[]{
\label{posthoc.precompetition.WithoutSeas.NoInputSel}
\begin{tabular}{|c|c|c|}
\hline
\multicolumn{3}{|c|}{\begin{minipage}{0.6\textwidth}\medskip \large Deseasonalization : {\bf Yes} - Input Selection : {\bf No} \end{minipage}} \\\hline
\textbf{WINNER} & \textbf{COMB} & \textbf{WCOMB} \\\hline

\begin{minipage}{.2\textwidth}
\begin{enumerate}[leftmargin=1em]
\item "DIRMO-SEL" (2.90) \\"DIRMO-WAVG" (3.21) \\"MIMO-ACFLIN" (3.21) \\"MIMO-LOO" (3.83) \\
\item "DIRMO-AVG" (4.74) \\"REC" (5.35) \\
\item "DIRREC" (6.33) \\"DIR" (6.42) \\
\end{enumerate}
\end{minipage} 
 &
\begin{minipage}{.2\textwidth}
\begin{enumerate}[leftmargin=1em]
\item "MIMO-LOO" (3.69) \\"DIRMO-SEL" (3.72) \\"DIRMO-WAVG" (3.72) \\"REC" (3.87) \\"MIMO-ACFLIN" (4.00) \\"DIRMO-AVG" (4.61) \\
\item "DIRREC" (6.15) \\"DIR" (6.23) \\
\end{enumerate}
\end{minipage} 
 &
\begin{minipage}{.2\textwidth}
\begin{enumerate}[leftmargin=1em]
\item "DIRMO-SEL" (3.66) \\"REC" (3.68) \\"DIRMO-WAVG" (3.79) \\"MIMO-LOO" (3.82) \\"MIMO-ACFLIN" (3.82) \\
\item "DIRMO-AVG" (4.77) \\
\item "DIRREC" (6.04) \\"DIR" (6.43) \\
\end{enumerate}
\end{minipage}\\ 
 
\hline 
\end{tabular}

}\\
\subfloat[]{
\label{posthoc.precompetition.WithoutSeas.InputSel}
\begin{tabular}{|c|c|c|}
\hline
\multicolumn{3}{|c|}{\begin{minipage}{0.6\textwidth}\medskip \large Deseasonalization : {\bf Yes} - Input Selection : {\bf Yes} \end{minipage}} \\\hline
\textbf{WINNER} & \textbf{COMB} & \textbf{WCOMB} \\\hline

\begin{minipage}{.2\textwidth}
\begin{enumerate}[leftmargin=1em]
\item "MIMO-ACFLIN" (1.76) \\
\item "MIMO-LOO" (2.59) \\"REC" (2.95) \\
\item "DIRREC" (3.77) \\"DIR" (3.95) \\
\end{enumerate}
\end{minipage} 
 &
\begin{minipage}{.2\textwidth}
\begin{enumerate}[leftmargin=1em]
\item "MIMO-ACFLIN" (2.38) \\"MIMO-LOO" (2.45) \\"REC" (2.57) \\
\item "DIRREC" (3.77) \\"DIR" (3.82) \\
\end{enumerate}
\end{minipage} 
 &
\begin{minipage}{.2\textwidth}
\begin{enumerate}[leftmargin=1em]
\item "MIMO-LOO" (2.34) \\"MIMO-ACFLIN" (2.34) \\"REC" (2.61) \\
\item "DIRREC" (3.74) \\"DIR" (3.97) \\
\end{enumerate}
\end{minipage}\\ 
 
\hline 
\end{tabular}

}
\caption{\emph{\textbf{Pre-competition phase}}. Group of strategies statistically significantly different (sorted by increasing ranking) provided by Friedman and post-hoc tests for the $12$ configurations. \label{posthoc.precompetition}}
\end{table}

\clearpage
The availability of the SMAPE* results, obtained according to the procedure used in \citep{amirnn5},  makes
possible the comparison of our pre-competition results with those of several others learning methods 
available in \citep{amirnn5}. For the sake of comparison, Table \ref{amirTable} reports the forecasting errors 
for some of the techniques considered in \citep{amirnn5}, notably Gaussian Process Regression~(GPR), Neural Network~(NN), Multiple Regression~(MULT-REGR), Simple Moving Average~(MOV-AVG), Holt's Exponential Smoothing and a combination (Combined) of such techniques. 
\begin{table}[!htp]
\centering
\begin{tabular}{|c|c|}
\hline
Model &SMAPE*\\
\hline
\hline
GPR-ITER & 19.90\\
\hline
GPR-DIR & 21.22\\
\hline
GPR-LEV & 20.19\\
\hline
NN-ITER & 21.11\\
\hline
NN-LEV & 19.83\\
\hline
MULT-REGR1 & 19.11\\
\hline
MULTI-REGR2 & 18.96\\
\hline
MULT-REGR3 & 18.94\\
\hline
MOV-AVG & 19.55\\
\hline
Holt's Exp Sm & 23.77\\
\hline
Combined & 18.95\\
\hline
\end{tabular}
\caption{Forecasting errors for the different forecasting models. \label{amirTable}}
\end{table}
The comparison shows that the best configuration of Table \ref{tableErrors.precompetition.WithoutSeas.InputSel}, that is the MIMO-ACFLIN strategy,  is competitive with all these models with a SMAPE* amounting to $18.81\%$.

\clearpage
\subsection{Competition results}

The SMAPE and ranking results for the \emph{competition} phase are presented in Table~\ref{tableErrors.competition} while the results of the post-hoc test are summarized in Table~\ref{posthoc.competition}.

%
\begin{table}[h!]
\setlength{\extrarowheight}{2pt}
\scriptsize
\begin{center}
\subfloat[]{
\label{tableErrors.competition.WithSeas.NoInputSel}
\begin{tabular}{|l||c|c||c|c||c|c|} 
\cline{2-7}
\multicolumn{1}{c}{} &\multicolumn{6}{|c|}{\large Deseasonalization : {\bf No} - Input Selection : {\bf No} }\\ 
\hline 
\multicolumn{1}{|l|}{\bf Strategies}&\multicolumn{2}{c||}{\bf WINNER}&\multicolumn{2}{c||}{\bf COMB}&\multicolumn{2}{c|}{\bf COMBW}\\ \cline{2-7}
\multicolumn{1}{|l|}{}&\multicolumn{1}{c|}{SMAPE* [Std err]}&\multicolumn{1}{c||}{Avg Rank}&\multicolumn{1}{c|}{SMAPE* [Std err]}&\multicolumn{1}{c||}{Avg Rank}&\multicolumn{1}{c|}{SMAPE* [Std err]}&\multicolumn{1}{c|}{Avg Rank}\\ \hline
DIR&24.48[0.52](7)&5.58(7)&23.06[0.50](6)&5.21(6)&22.89[0.48](6)&5.20(7)\\ 
REC&24.22[0.62](6)&4.96(6)&23.71[0.52](7)&5.24(7)&23.54[0.54](7)&5.05(6)\\ 
DIRREC&44.97[0.85](8)&7.93(8)&40.37[0.80](8)&7.97(8)&38.17[0.72](8)&7.95(8)\\ 
\hline
MIMO-LOO&{\bf21.92}[0.56](1)&{\bf2.90}(1)&{\bf21.55}[0.50](1)&{\bf2.84}(1)&{\bf21.64}[0.49](1)&{\bf2.79}(1)\\ 
MIMO-ACFLIN&22.45[0.54](3)&3.65(3)&21.57[0.50](2)&3.08(2)&21.64[0.49](2)&2.79(2)\\ 
DIRMO-SEL&22.60[0.54](5)&3.99(5)&22.27[0.49](5)&4.46(5)&22.50[0.50](5)&4.84(5)\\ 
DIRMO-AVG&22.55[0.53](4)&3.78(4)&21.73[0.49](4)&3.68(4)&21.93[0.49](4)&3.84(4)\\ 
DIRMO-WAVG&22.36[0.54](2)&3.21(2)&21.70[0.49](3)&3.52(3)&21.87[0.49](3)&3.55(3)\\ 
\hline
\end{tabular}

}\\
\subfloat[]{
\label{tableErrors.competition.WithSeas.InputSel}
\begin{tabular}{|l||c|c||c|c||c|c|} 
\cline{2-7}
\multicolumn{1}{c}{} &\multicolumn{6}{|c|}{\large Deseasonalization : {\bf No} - Input Selection : {\bf Yes} }\\ 
\hline 
\multicolumn{1}{|l|}{\bf Strategies}&\multicolumn{2}{c||}{\bf WINNER}&\multicolumn{2}{c||}{\bf COMB}&\multicolumn{2}{c|}{\bf COMBW}\\ \cline{2-7}
\multicolumn{1}{|l|}{}&\multicolumn{1}{c|}{SMAPE* [Std err]}&\multicolumn{1}{c||}{Avg Rank}&\multicolumn{1}{c|}{SMAPE* [Std err]}&\multicolumn{1}{c||}{Avg Rank}&\multicolumn{1}{c|}{SMAPE* [Std err]}&\multicolumn{1}{c|}{Avg Rank}\\ \hline
DIR&24.43[0.52](4)&3.28(4)&23.14[0.48](4)&3.22(4)&23.25[0.48](4)&3.24(4)\\ 
REC&22.73[0.69](3)&2.28(2)&22.45[0.67](3)&2.53(3)&22.57[0.65](3)&2.42(3)\\ 
DIRREC&44.91[0.86](5)&4.94(5)&39.88[0.77](5)&4.93(5)&38.04[0.71](5)&4.95(5)\\ 
\hline
MIMO-LOO&{\bf22.18}[0.53](1)&{\bf2.12}(1)&{\bf21.78}[0.49](1)&{\bf2.09}(1)&{\bf21.84}[0.49](1)&{\bf2.19}(1)\\ 
MIMO-ACFLIN&22.62[0.51](2)&2.39(3)&21.83[0.50](2)&2.24(2)&21.84[0.49](2)&2.19(2)\\ 
\hline
\end{tabular}

}\\
\subfloat[]{
\label{tableErrors.competition.WithoutSeas.NoInputSel}
\begin{tabular}{|l||c|c||c|c||c|c|} 
\cline{2-7}
\multicolumn{1}{c}{} &\multicolumn{6}{|c|}{\large Deseasonalization : {\bf Yes} - Input Selection : {\bf No} }\\ 
\hline 
\multicolumn{1}{|l|}{\bf Strategies}&\multicolumn{2}{c||}{\bf WINNER}&\multicolumn{2}{c||}{\bf COMB}&\multicolumn{2}{c|}{\bf COMBW}\\ \cline{2-7}
\multicolumn{1}{|l|}{}&\multicolumn{1}{c|}{SMAPE* [Std err]}&\multicolumn{1}{c||}{Avg Rank}&\multicolumn{1}{c|}{SMAPE* [Std err]}&\multicolumn{1}{c||}{Avg Rank}&\multicolumn{1}{c|}{SMAPE* [Std err]}&\multicolumn{1}{c|}{Avg Rank}\\ \hline
DIR&23.98[0.55](8)&6.36(8)&21.65[0.46](8)&6.05(8)&21.75[0.47](7)&6.14(8)\\ 
REC&23.12[0.59](6)&5.47(6)&21.39[0.49](6)&5.04(6)&21.86[0.49](8)&5.40(6)\\ 
DIRREC&23.51[0.60](7)&6.15(7)&21.58[0.52](7)&5.78(7)&21.57[0.51](6)&5.67(7)\\ 
\hline
MIMO-LOO&21.11[0.54](4)&3.69(4)&20.27[0.47](4)&3.77(3)&20.34[0.47](3)&3.69(3)\\ 
MIMO-ACFLIN&{\bf20.25}[0.47](1)&{\bf3.05}(1)&20.18[0.46](2)&3.67(2)&20.34[0.47](4)&3.69(4)\\ 
DIRMO-SEL&20.85[0.51](2)&3.45(3)&20.18[0.46](3)&3.77(4)&{\bf20.23}[0.45](1)&{\bf3.45}(1)\\ 
DIRMO-AVG&21.66[0.51](5)&4.47(5)&20.38[0.46](5)&4.34(5)&20.48[0.46](5)&4.38(5)\\ 
DIRMO-WAVG&20.97[0.51](3)&3.35(2)&{\bf20.15}[0.46](1)&{\bf3.59}(1)&20.23[0.46](2)&3.59(2)\\ 
\hline
\end{tabular}

}\\
\subfloat[]{
\label{tableErrors.competition.WithoutSeas.InputSel}
\begin{tabular}{|l||c|c||c|c||c|c|} 
\cline{2-7}
\multicolumn{1}{c}{} &\multicolumn{6}{|c|}{\large Deseasonalization : {\bf Yes} - Input Selection : {\bf Yes} }\\ 
\hline 
\multicolumn{1}{|l|}{\bf Strategies}&\multicolumn{2}{c||}{\bf WINNER}&\multicolumn{2}{c||}{\bf COMB}&\multicolumn{2}{c|}{\bf COMBW}\\ \cline{2-7}
\multicolumn{1}{|l|}{}&\multicolumn{1}{c|}{SMAPE* [Std err]}&\multicolumn{1}{c||}{Avg Rank}&\multicolumn{1}{c|}{SMAPE* [Std err]}&\multicolumn{1}{c||}{Avg Rank}&\multicolumn{1}{c|}{SMAPE* [Std err]}&\multicolumn{1}{c|}{Avg Rank}\\ \hline
DIR&23.91[0.50](5)&4.14(5)&21.54[0.47](5)&3.94(5)&21.58[0.48](5)&3.91(5)\\ 
REC&22.57[0.64](3)&3.05(3)&21.39[0.63](3)&2.96(3)&21.45[0.64](3)&2.93(3)\\ 
DIRREC&23.66[0.58](4)&3.79(4)&21.48[0.52](4)&3.50(4)&21.47[0.51](4)&3.50(4)\\ 
\hline
MIMO-LOO&20.74[0.51](2)&2.14(2)&{\bf20.28}[0.53](1)&{\bf2.27}(1)&{\bf20.28}[0.52](1)&{\bf2.33}(1)\\ 
MIMO-ACFLIN&{\bf20.39}[0.54](1)&{\bf1.87}(1)&20.28[0.53](2)&2.32(2)&20.28[0.52](2)&2.33(2)\\ 
\hline
\end{tabular}

}
\end{center}
\caption{\emph{\textbf{Competition phase}}. Average forecasting errors (SMAPE*) with average ranking for each strategy in the 12 different configurations. The numbers in round bracket represent the ranking within the column. \label{tableErrors.competition}}
\end{table}

\begin{table}[!htp]
\centering\tiny
\subfloat[]{
\label{posthoc.competition.WithSeas.NoInputSel}
\begin{tabular}{|c|c|c|}
\hline
\multicolumn{3}{|c|}{\begin{minipage}{0.6\textwidth}\medskip \large Deseasonalization : {\bf No} - Input Selection : {\bf No} \end{minipage}} \\\hline
\textbf{WINNER} & \textbf{COMB} & \textbf{WCOMB} \\\hline

\begin{minipage}{.2\textwidth}
\begin{enumerate}[leftmargin=1em]
\item "MIMO-LOO" (2.90) \\"DIRMO-WAVG" (3.21) \\"MIMO-ACFLIN" (3.65) \\"DIRMO-AVG" (3.78) \\"DIRMO-SEL" (3.99) \\
\item "REC" (4.96) \\"DIR" (5.58) \\
\item "DIRREC" (7.93) \\
\end{enumerate}
\end{minipage} 
 &
\begin{minipage}{.2\textwidth}
\begin{enumerate}[leftmargin=1em]
\item "MIMO-LOO" (2.84) \\"MIMO-ACFLIN" (3.08) \\"DIRMO-WAVG" (3.52) \\"DIRMO-AVG" (3.68) \\"DIRMO-SEL" (4.46) \\"DIR" (5.21) \\"REC" (5.24) \\
\item "DIRREC" (7.97) \\
\end{enumerate}
\end{minipage} 
 &
\begin{minipage}{.2\textwidth}
\begin{enumerate}[leftmargin=1em]
\item "MIMO-LOO" (2.79) \\"MIMO-ACFLIN" (2.79) \\"DIRMO-WAVG" (3.55) \\"DIRMO-AVG" (3.84) \\
\item "DIRMO-SEL" (4.84) \\"REC" (5.05) \\"DIR" (5.20) \\
\item "DIRREC" (7.95) \\
\end{enumerate}
\end{minipage}\\ 
 
\hline 
\end{tabular}

}\\
\subfloat[]{
\label{posthoc.competition.WithSeas.InputSel}
\begin{tabular}{|c|c|c|}
\hline
\multicolumn{3}{|c|}{\begin{minipage}{0.6\textwidth}\medskip \large Deseasonalization : {\bf No} - Input Selection : {\bf Yes} \end{minipage}} \\\hline
\textbf{WINNER} & \textbf{COMB} & \textbf{WCOMB} \\\hline

\begin{minipage}{.2\textwidth}
\begin{enumerate}[leftmargin=1em]
\item "MIMO-LOO" (2.12) \\"REC" (2.28) \\"MIMO-ACFLIN" (2.39) \\
\item "DIR" (3.28) \\
\item "DIRREC" (4.94) \\
\end{enumerate}
\end{minipage} 
 &
\begin{minipage}{.2\textwidth}
\begin{enumerate}[leftmargin=1em]
\item "MIMO-LOO" (2.09) \\"MIMO-ACFLIN" (2.24) \\"REC" (2.53) \\
\item "DIR" (3.22) \\
\item "DIRREC" (4.93) \\
\end{enumerate}
\end{minipage} 
 &
\begin{minipage}{.2\textwidth}
\begin{enumerate}[leftmargin=1em]
\item "MIMO-LOO" (2.19) \\"MIMO-ACFLIN" (2.19) \\"REC" (2.42) \\
\item "DIR" (3.24) \\
\item "DIRREC" (4.95) \\
\end{enumerate}
\end{minipage}\\ 
 
\hline 
\end{tabular}

}\\
\subfloat[]{
\label{posthoc.competition.WithoutSeas.NoInputSel}
\begin{tabular}{|c|c|c|}
\hline
\multicolumn{3}{|c|}{\begin{minipage}{0.6\textwidth}\medskip \large Deseasonalization : {\bf Yes} - Input Selection : {\bf No} \end{minipage}} \\\hline
\textbf{WINNER} & \textbf{COMB} & \textbf{WCOMB} \\\hline

\begin{minipage}{.2\textwidth}
\begin{enumerate}[leftmargin=1em]
\item "MIMO-ACFLIN" (3.05) \\"DIRMO-WAVG" (3.35) \\"DIRMO-SEL" (3.45) \\"MIMO-LOO" (3.69) \\"DIRMO-AVG" (4.47) \\
\item "REC" (5.47) \\"DIRREC" (6.15) \\"DIR" (6.36) \\
\end{enumerate}
\end{minipage} 
 &
\begin{minipage}{.2\textwidth}
\begin{enumerate}[leftmargin=1em]
\item "DIRMO-WAVG" (3.59) \\"MIMO-ACFLIN" (3.67) \\"MIMO-LOO" (3.77) \\"DIRMO-SEL" (3.77) \\"DIRMO-AVG" (4.34) \\"REC" (5.04) \\"DIRREC" (5.78) \\"DIR" (6.05) \\
\end{enumerate}
\end{minipage} 
 &
\begin{minipage}{.2\textwidth}
\begin{enumerate}[leftmargin=1em]
\item "DIRMO-SEL" (3.45) \\"DIRMO-WAVG" (3.59) \\"MIMO-LOO" (3.69) \\"MIMO-ACFLIN" (3.69) \\"DIRMO-AVG" (4.38) \\
\item "REC" (5.40) \\"DIRREC" (5.67) \\"DIR" (6.14) \\
\end{enumerate}
\end{minipage}\\ 
 
\hline 
\end{tabular}

}\\
\subfloat[]{
\label{posthoc.competition.WithoutSeas.InputSel}
\begin{tabular}{|c|c|c|}
\hline
\multicolumn{3}{|c|}{\begin{minipage}{0.6\textwidth}\medskip \large Deseasonalization : {\bf Yes} - Input Selection : {\bf Yes} \end{minipage}} \\\hline
\textbf{WINNER} & \textbf{COMB} & \textbf{WCOMB} \\\hline

\begin{minipage}{.2\textwidth}
\begin{enumerate}[leftmargin=1em]
\item "MIMO-ACFLIN" (1.87) \\"MIMO-LOO" (2.14) \\
\item "REC" (3.05) \\
\item "DIRREC" (3.79) \\"DIR" (4.14) \\
\end{enumerate}
\end{minipage} 
 &
\begin{minipage}{.2\textwidth}
\begin{enumerate}[leftmargin=1em]
\item "MIMO-LOO" (2.27) \\"MIMO-ACFLIN" (2.32) \\
\item "REC" (2.96) \\
\item "DIRREC" (3.50) \\"DIR" (3.94) \\
\end{enumerate}
\end{minipage} 
 &
\begin{minipage}{.2\textwidth}
\begin{enumerate}[leftmargin=1em]
\item "MIMO-LOO" (2.33) \\"MIMO-ACFLIN" (2.33) \\
\item "REC" (2.93) \\
\item "DIRREC" (3.50) \\"DIR" (3.91) \\
\end{enumerate}
\end{minipage}\\ 
 
\hline 
\end{tabular}

}
\caption{\emph{\textbf{Competition phase}}. Group of strategies statistically significantly different (sorted by increasing ranking) provided by Friedman and post-hoc tests for the $12$ configurations. \label{posthoc.competition}}
\end{table}

\clearpage

The pre-competition results presented in the previous section suggest us to use the MIMO-ACFLIN strategy with the \verb+Comb+ model selection approach by removing the seasonality and applying the input selection procedure, since this configuration obtains the smallest forecasting error~($18.81 \%$). 

\begin{figure}
\centering
\vspace{-0.3cm}
\includegraphics[trim = 0cm 0.5cm 0cm 1.9cm, clip,scale=0.8,scale=0.6]{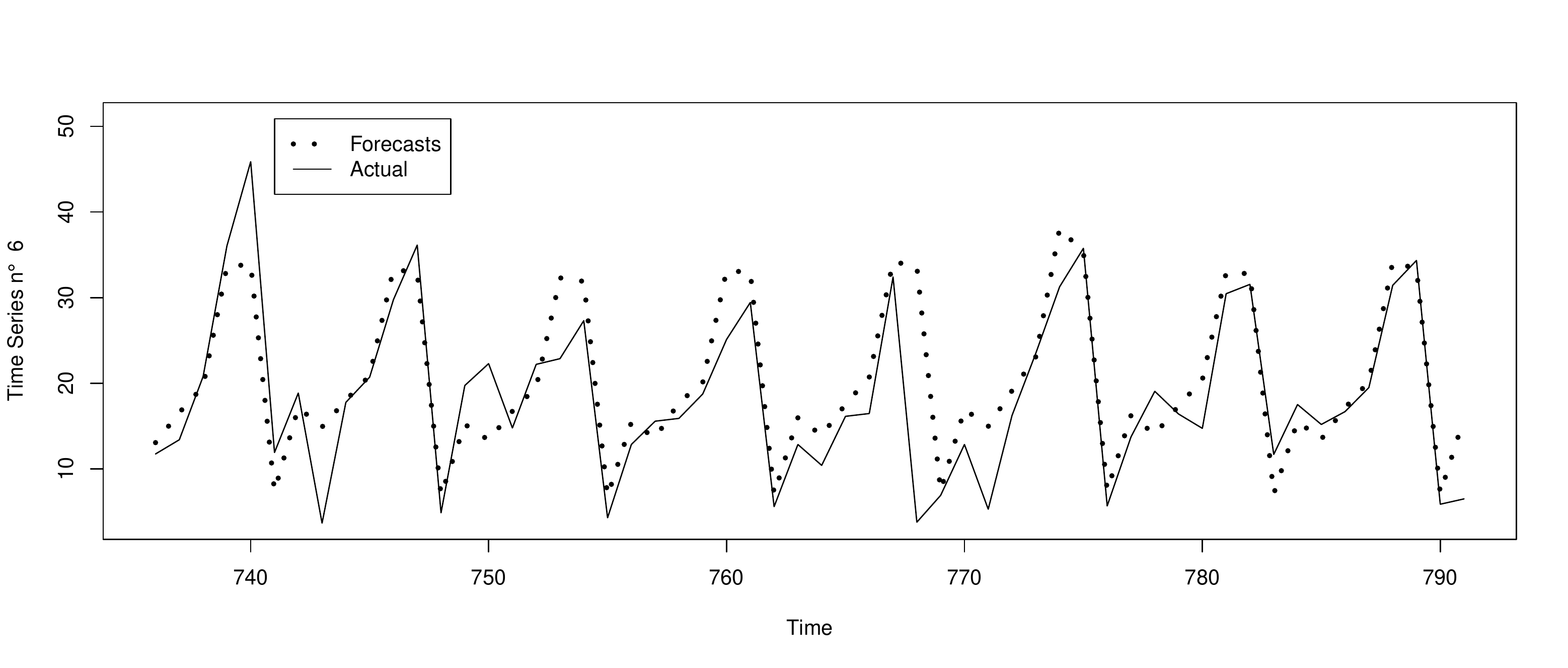} \\
\includegraphics[trim = 0cm 0.5cm 0cm 1.9cm, clip,scale=0.8,scale=0.6]{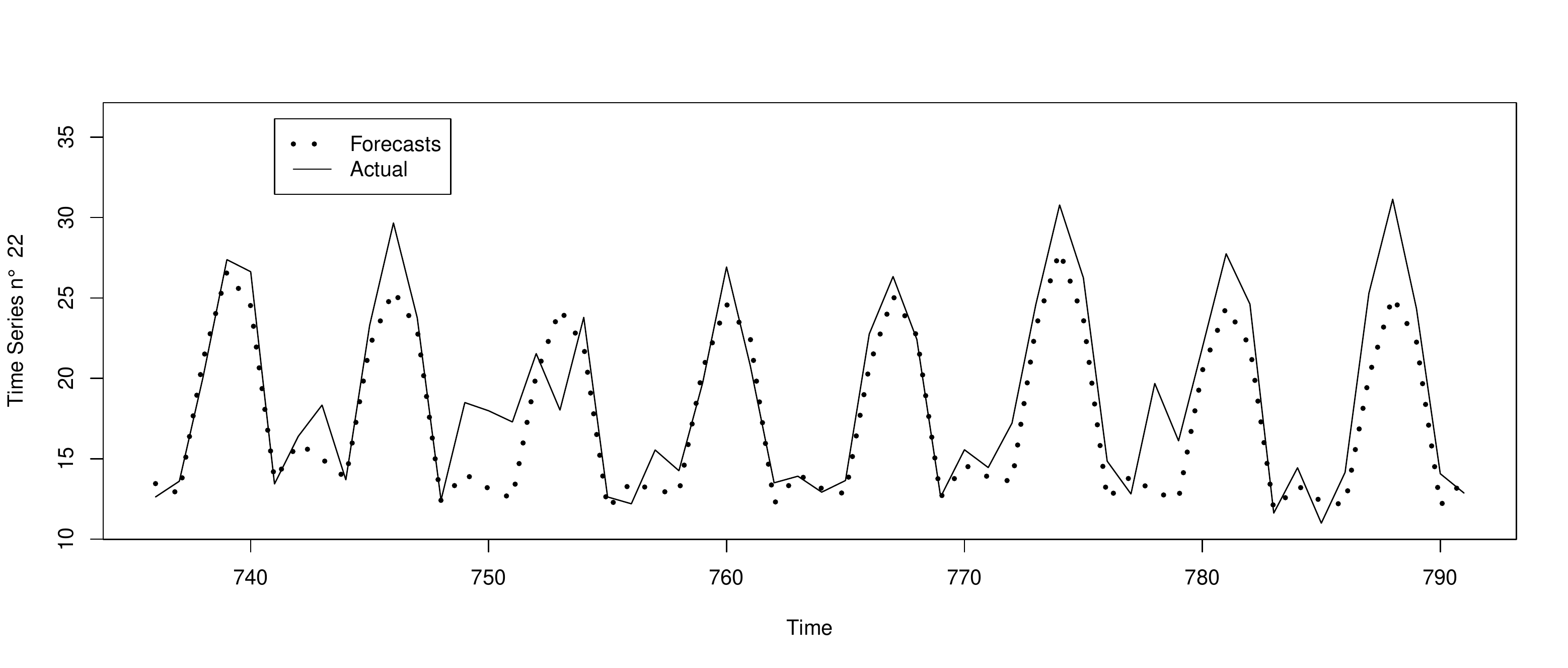} \\
\includegraphics[trim = 0cm 0.5cm 0cm 1.9cm, clip,scale=0.8,scale=0.6]{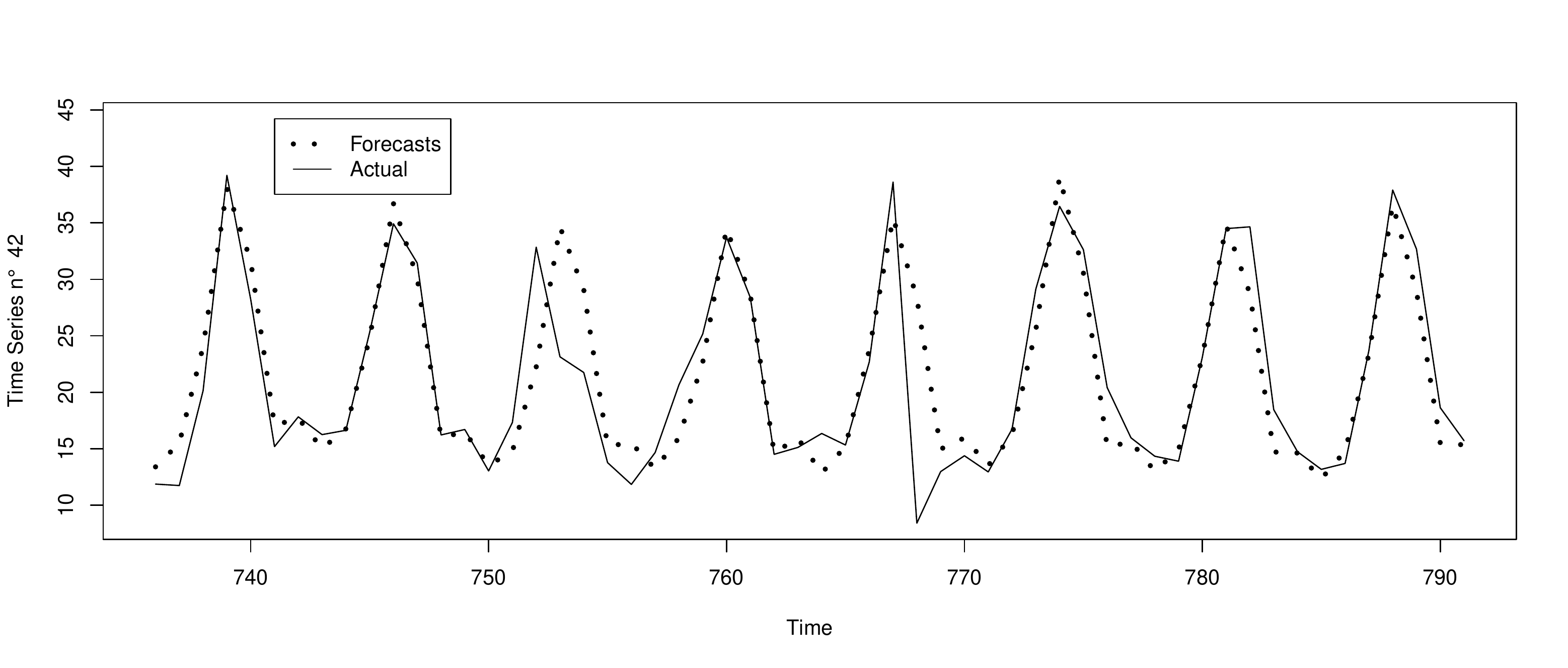} \\
\includegraphics[trim = 0cm 0.5cm 0cm 1.9cm, clip,scale=0.8,scale=0.6]{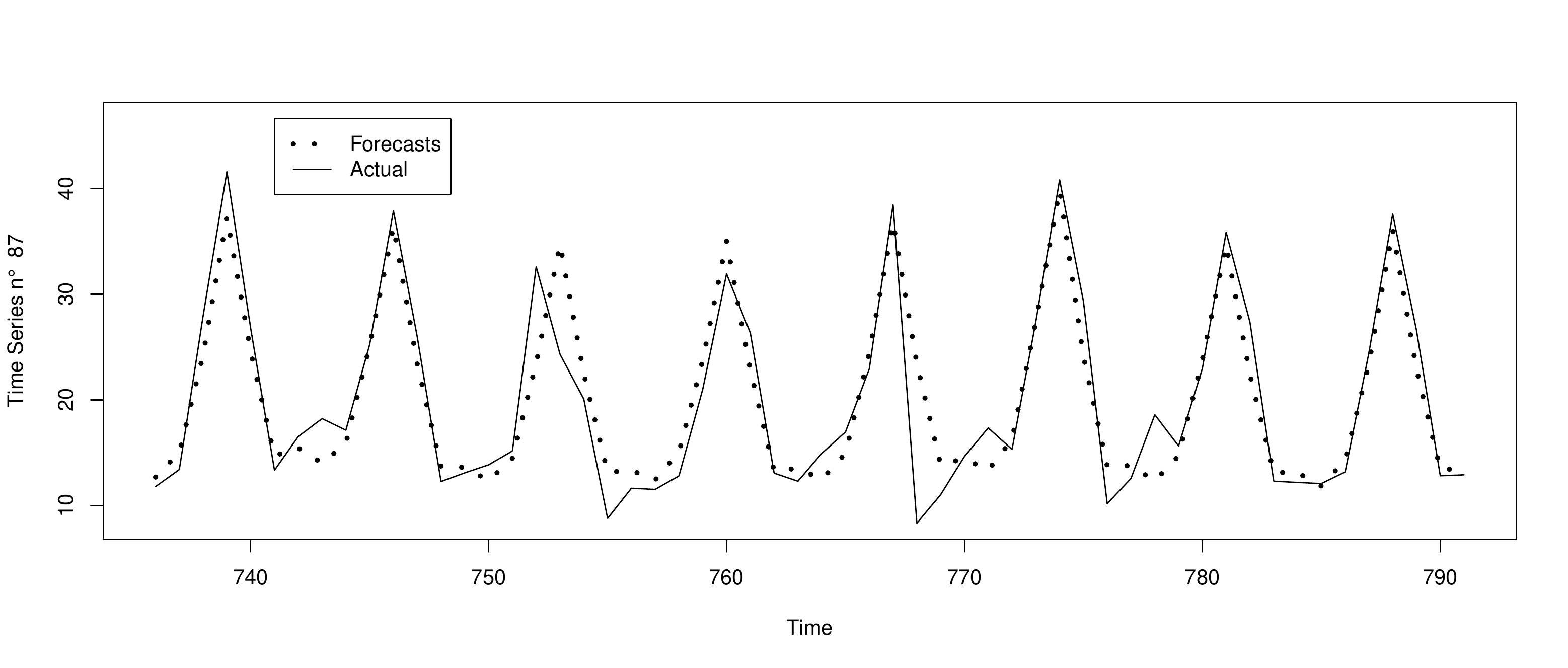} \\
\caption{ The forecast versus the actual of the MIMO-ACFLIN strategy for four NN5 time series. \label{NN5TimeSeriesForecasts}}
\end{figure}

By using the MIMO-ACFLIN strategy and the corresponding configuration in the competition phase, we would generate forecasts with a SMAPE* equals to $20.28 \%$ which is quite good compared to the best computational intelligence entries of the competition as shown in Table \ref{NN5results}. Figure \ref{NN5TimeSeriesForecasts} shows the forecasts of the MIMO-ACFLIN strategy versus the actual values for four NN5 time series to illustrate the forecasting capability of this strategy.

\begin{table}[!htp]
\centering
\begin{tabular}{|c|c|}
\hline
Model &SMAPE*\\
\hline
\hline
\textbf{MIMO-ACFLIN} & $\mathbf{20.28}$\\
\hline
Andrawis & 20.4\\
\hline
Vogel & 20.5\\
\hline
D'yakonov & 20.6\\
\hline
Rauch & 21.7\\
\hline
Luna & 21.8\\
\hline
Wichard & 22.1\\
\hline
Gao & 22.3\\
\hline
Puma-Villanueva & 23.7\\
\hline
Dang & 23.77\\
\hline
Pasero & 25.3\\
\hline
\end{tabular}
\caption{Forecasting errors for different computational intelligence forecasting models which participate to the NN5 forecasting competition. \label{NN5results}}
\end{table}

\clearpage

\clearpage

\subsection{Discussion}

From all presented results one can deduce the following observations below.
These findings refer mainly to the pre-competition results. But, one can easily see
that they mostly also apply to the competition phase results.

\begin{itemize}

\item The overall best method is MIMO-ACFLIN, used with input selection, deseasonalization
and equal weight combination (COMB).

\item The Multiple-Output strategies (MIMO and DIRMO) are invariably the best strategies.
They beat the Single-Output strategies, such as DIR, REC, and DIRREC.
Both MIMO and DIRMO give comparable performancce. 
For DIRMO, the selection of the parameter $s$ is critical, since it has a great impact on the performance. 
Should there be an improved selection approach, this strategy would have a big potential. 

\item
Both versions of MIMO are comparable. Also the versions of
DIRMO give close results, with perhaps DIRMO-WAVG a little better
than the other two versions.

\item Among the Single-Output strategies, the REC strategy has almost always a smaller SMAPE and a better
ranking than the DIR strategy. DIRREC is the worse strategy overall,
and gives especially low accuracy when no deseasonalization is performed.

\item Deseasonalization leads to consistently better results (in 38 out of 39 models).
This result is consistent with some other studies, such as \citep{Zhang2005501}.
The possible reason for this is that when no deseasonalization is performed, we are putting
a higher burden on the model to forecast the
future seasonal pattern plus the trend and the other aspects,
which apparently is hard to simultaneously satisfy.

\item Input selection is especially beneficial when we perform
a deseasonalization. Absent deseasonalization, the results are mixed
(as to whether input selection improves the results or not).
The possible explanation is that when no deseasonalization is performed,
the model needs all the previous cycle to construct the future seasonal
pattern. Performing an input selection will deprive it from essential information. 

\item Concerning the model selection aspect, both combination
approaches (COMB and WCOMB) are superior to the winner-take-all (WINNER).
Both COMB and WCOMB are comparable, and the results do not
differ by much. This is consistent with much of the findings 
in forecast combination literature, e.g.~\citep{amirnn5,Clemen89,forecastcomb,Andrawis2010}

\item The relative performance and ranking of the different strategies
 is persistent. Most findings that are based on the pre-competition 
 results are true for the competition phase results. This is also true
 for the findings concerning the deseasonalization, input selection, and model selection.
 This persistence is reassuring, as we can have some confidence
 in relying on the test or validation results for selecting the best strategies.
 
 \item The best strategy based on the pre-competition data, the MIMO-ACFLIN method,
 would have topped all computational intelligence entries of the NN5 competition in the true competition
 hold-out data.
 
\end{itemize}

\section{Conclusion}
\label{sectionConclusion}
Forecasting a time series many steps into the future is a very
hard problem because the larger the forecast horizon, the higher is the uncertainty.
In this paper we presented a comparative review of existing strategies for multi-step ahead forecasting, together with an extensive comparison, applied on the 111 time series of the NN5 forecasting competition.
The comparison gave some interesting lessons that could help researchers channel their
experiments into the most promising approaches.
The most consistent findings are that Multiple-Output approaches are invariably better than Single-Output
approaches. Also, deseasonalization had a very considerable positive impact on the
performance. Finally, the results are clearly quite persistent.
So, selecting the best strategy based on testing performance is a very potent approach.
A possible direction for future research could therefore be 
developing other new improved Multiple-Output strategies. Also, possibly tailoring 
deseasonalization methods specifically for Multiple-Output strategies could also be a promising
research point.
 
\section*{Acknowledgments}

We would like to thanks the authors of the paper \citep{extensionDemsar} for making their methods available at \url{http://sci2s.ugr.es/keel/multipleTest.zip}.

\bibliographystyle{plainnat}
\bibliography{ijf,mybib}

\end{document}